\documentclass{article}
\usepackage[top=1.5in, bottom=1.5in, left=1.25in, right=1.25in]{geometry}

\usepackage{latexopts}

\usepackage[compatibility=false]{caption}
\usepackage{subcaption}

\newcommand{\TV}{\operatorname{TV}}
\newcommand{\excess}{\operatorname{excess}}

\usepackage{natbib}

\begin{document}

\title{An Algorithmic Theory of Dependent Regularizers\\Part 1: Submodular Structure}

\author{
  Hoyt Koepke \\
  Department of Statistics\\
  University of Washington\\
  Seattle, WA 98105 \\
  \texttt{hoytak@stat.washington.edu} \\
  \\
  Marina Meila \\
  Department of Statistics\\
  University of Washington \\
  Seattle, WA 98105 \\
  \texttt{mmp@stat.washington.edu}
}

\maketitle

\begin{abstract}
  We present an exploration of the rich theoretical connections
  between several classes of regularized models, network flows, and
  recent results in submodular function theory.  This work unifies key
  aspects of these problems under a common theory, leading to novel
  methods for working with several important models of interest in
  statistics, machine learning and computer vision. 

  In Part 1, we review the concepts of network flows and submodular
  function optimization theory foundational to our results. We then
  examine the connections between network flows and the minimum-norm
  algorithm from submodular optimization, extending and improving
  several current results. This leads to a concise representation of
  the structure of a large class of pairwise regularized models
  important in machine learning, statistics and computer vision.

  In Part 2, we describe the full regularization path of a class of
  penalized regression problems with dependent variables that includes
  the graph-guided LASSO and total variation constrained models.  This
  description also motivates a practical algorithm. This allows us to
  efficiently find the regularization path of the discretized version
  of TV penalized models.  Ultimately, our new algorithms scale up to
  high-dimensional problems with millions of variables.
\end{abstract}

\newpage

\section{Introduction}
\label{sec:ch:background}

High-dimensional data is a central focus of modern research in
statistics and machine learning.  Recent technological advances in a
variety of scientific fields for gathering and generating data,
matched by rapidly increasing computing power for analysis, has
attracted significant research into the statistical questions
surrounding structured data. Numerous models and computational
techniques have emerged recently to work with these data types.

Our primary contribution is a theoretical framework enabling the
efficient optimization of models that work with high-dimensional
models in which the predictors are believed to be dependent,
correlated, or sparse.  In particular, we propose a new approach for
estimation of certain types of dependent predictors, in which the
model incorporates a prior belief that many of the predictors take
similar or identical values.  This has been a hot topic of research in
recent years, with applications in genomics, image analysis, graphical
models, and several other areas.  However, efficient estimation
involving structured and dependent predictors has proven to be quite
challenging. Our main contribution, which includes a number of related
theoretical results, is a theoretical framework and algorithmic
approach that unlocks a large and particularly thorny class of these
models.

After laying out the context for our research in the next section, we
outline our main results as well as the needed background in section
\ref{sec:b:outline}.  Our contributions build on recent results in
optimization theory and combinatorial optimization, which we describe
in detail.  In \ref{sec:ch:structure}, we present a clear theoretical
connection between submodular optimization theory, network flows, and
the proximal operator in regularized models such as the graph-guided
LASSO.  Then, in \ref{sec:ch:general}, we extend the underlying
submodular optimization theory to allow a weighted version of the
underlying submodular problem.  In the context of regularized
regression, this allows the use of additional unary terms in the
regularizer.   

\section{Overview of Statistical Models and Regularization}
\label{sec:b:models}

To begin, consider a simple regression model, in which we have $N$
observations $\my = (\m y_1, \my_2, ..., \m y_N)$ and $n$ predictors $\m u =
(u_1, u_2, ..., u_n)$, with
\begin{equation}
  \my_i = \mA\m u + \meps_i, \qquad i = 1,2,...,N
  \label{eq:lZ3AS}
\end{equation}
where $\meps_1, \meps_2, ..., \meps_N$ are independent random noise
vectors with $\E \meps_i = \m0$; typically, these are assumed to be
i.i.d. Gaussian.  In many high dimensional contexts, the estimation of
$\mhu$ may be problematic using classical methods.  For example, $n$
may be much larger than $N$, making the problem ill-posed as many
possible values of $\m u$ map to the same response. ($n$ and $N$ are
used here instead of the more common $p$ and $n$ to be consistent with
the optimization literature we connect this problem to.)
Additionally, many predictors of interest may have a negligible or
nearly identical affect on $\my$; eliminating or grouping these
predictors is then desirable.  To handle this, a number of
sophisticated approaches have been proposed to incorporate variable
selection or aggregation into the statistical estimation problem.

One common and well-studied approach is to add a penalty term, or
regularizer, to the log-likelihood that enforces some prior belief
about the structure of $\m u$ \citep{bickel2006regularization,
  hastie2009linear}.  In this setup, estimating the predictor $\m u$
involves finding the minimizer of a log-likelihood term or loss
function $\Lcal$ plus a regularization term $\Phi$:
\begin{equation}
  \mhu = \argmin_{\m u \in \Reals^n} \Lcal(\m u, \my) + \lambda\Phi(\m u),
  \label{eq:oqqvN}
\end{equation}
where $\lambda$ controls the strength of the regularization $\lambda
\Phi(\m u)$.  

Many well studied and frequently used models fall into this context.
For example, if $\Lcal$ is the log-likelihood from a multivariate
Gaussian distribution, one of the oldest regularization techniques is
the $L_2$-norm, which gives us the classic technique of ridge
regression \citep{hoerl1970ridge}, where
\begin{equation}
  \mhu_{\text{ridge}} = \argmin_{\m u \in \Reals^n} \sqnorm{\my - \mA \m u} 
  + \lambda \sqnorm{\m u}.
  \label{eq:410Rx}
\end{equation}
In this example, the regularization term is typically used as an
effective way of dealing with an ill-conditioned inverse problem in
the least squares context or as a simple way of preventing $\m u$ from
over-fitting the model \citep{bishop2006pra}.

In the Bayesian context, \eqref{eq:oqqvN} often corresponds directly
to finding the maximum a posteriori estimate in the classic
likelihood-prior formulation, i.e. 
\begin{align}
  p(\m u \C \m y) &\propto p(\m y \C \m u) p(\m u \; ; \; \lambda, \theta) \\
  &\propto e^{-\Lcal(\m u, \my)} \times e^{-\lambda \Phi_\theta(\m u)}. 
  \label{eq:RUAZq}
\end{align}
where $\lambda$ and $\theta$ are hyperparameters controlling the
behavior of the prior distribution.  In this formulation, the prior
captures the belief encoded by the regularization term.  For example,
the ridge regression problem of \eqref{eq:410Rx} corresponds to using
a standard multivariate Gaussian likelihood but assumes a $0$-mean
spherical Gaussian prior distribution over $\m u$.

More recently, the Least Absolute Shrinkage and Selection Operator
(LASSO), is used frequently to promote sparsity in the resulting
estimator \citep{tibshirani1996regression, hastie2005elements,
  hastie2009linear}.  This approach uses the $L_1$-norm as the
regularization term, i.e.
\begin{equation}
  \mhu_{\text{lasso}} = \argmin_{\m u \in \Reals^n} \sqnorm{\my - \mA \m u} + \lambda \lnorm{\m u}{1}
  \label{eq:420Rx}
\end{equation}
The LASSO problem has been analyzed in detail as a method for
simultaneous estimation and variable selection, as the $L_1$ penalty
tends to set many of the predictors to $0$.  The consistency and
theoretical properties of this model, for both estimation and variable
selection, are well studied \citep{meinshausen2009lasso,
  bunea2007sparsity, bunea2006aggregation, van2008high, van2007non,
  bickel2006regularization}.  Many variants of this problem have also
been proposed and analyzed.  These include the elastic net, in which
$\Phi(\m u) = \alpha \lnorm{\m u}{1} + (1 - \alpha) \sqnorm{\m u}$,
with $\alpha \in \Icc{0,1}$; this generalizes both ridge regression
and LASSO \citep{zou2005regularization}.  Data-adaptive versions of
these estimators are analyzed in \citep{zou2006adaptive,
  zou2005regularization, zou2009adaptive}.  In addition, there are
many efficient algorithms to quickly solve these problems, which is
one of the reasons they are used frequently in practice
\citep{friedman2008regularization, friedman2010regularization,
  friedman2007pathwise, efron2004least, beck2009fast}.

\subsection{Graph Structured Dependencies}
\label{sec:b:dependencies}

While there are other possible regularization strategies on individual
parameters, of most interest to us is the incorporation of correlation
structures and dependencies among the predictors into the prior
distribution or regularization term.  In many cases, this translates
into penalizing the differences of predictors.  This has recently been
attracting significant algorithmic and theoretical interest as a way
to effectively handle dependency structures in the data.

In our particular context, we are interested in the MAP estimate of
models with Markov Random Field prior.  The resulting log-linear
models consist of a collection of unary and pairwise terms that
capture the dependency structure of the problem.  In particular, the
priors we are interested in enforce similarity between neighboring
variables, where ``neighbor'' is defined according to the graph
structure of the MRF prior.  The type of dependency we are looking at
is determined by the 

The simplest form of model dependency in the parameters is captured by
the {\it Fused LASSO} problem \citep{tibshirani2005sparsity}, which
penalizes pairwise differences between terms in an ordered problem to
be
\begin{equation}
  \m u^*_{\text{fused}} = \argmin_{\m u \in \Reals^n} = \sqnorm{\mA \m u - \my}
  + \lambda \Tbr{w_1\lnormof{\m u}{1} +  w_2 \sum_{i = 1}^{n-1} \absof{u_{i+1} - u_{i}}},
  \label{eq:r9Csf}
\end{equation}
where $w_1,w_2 \geq 0$ control the balance between the $L_1$-norm
controlling variable sparsity and the sum of ordered differences that
effectively penalizes changepoints. This problem has gained some
recent attention in the statistics community in the context of
non-parametric regression \citep{dumbgen2009extensions,
  cho2011multiscale, davies2008approximating}, group sparsity
\citep{tibshirani2005sparsity, bleakley2011group}, and change-point
detection \citep{bleakley2011group}.  From the optimization
perspective, several algorithms to find the solution to this problem
have been proposed; we refer the reader to \citet{liu2010efficient,
  ye2011split,bach2012structured} or \citet{friedman2007pathwise} for
discussions of this particular problem.

The generalized problem, in which the pairwise terms are not required
to be ordered, is of significant practical interest in both the
statistics and machine learning communities.  Here, the pairwise
interactions are controlled with an arbitrary graph of weighted
difference penalties:
\begin{equation}
  \m u^*_{\text{graph}} = \argmin_{\m u \in \Reals^n} = \sqnorm{\my - \mA \m u}
  + \lambda \Tbr{w_1\lnormof{\m u}{1} +  \sum_{1 \leq i < j \leq n} w_{ij} \absof{u_i - u_j}},
  \label{eq:r9C33}
\end{equation}
where $w_1 \geq 0$ controls the sparsity of the individual predictors,
and $w_{ij} \geq 0$ penalizes differences, typically between
predictors known to be correlated.  This problem has become known as
the {\it graph guided LASSO} \cite{NIPS2013_4934}. 

This model -- often with $w_1 = 0$ -- arises in the context of
``roughness'' penalized regression in which differences between
neighboring regions are penalized
\citep{belkin2004regularization}. \citet*{besag1995bayesian} examined
pairwise interactions on Markov random fields, which describe a prior
distribution on the edges of an undirected graph; the proposed models
were then solved with MCMC methods. Similar problems also arise in the
estimation of sparse covariance matrices as outlined by
\citet*{friedman2008sparse}; there, structure learning is the goal and
the pairwise interaction terms are not formed explicitly.

Similarly, \citet*{kovac2011nonparametric} analyzes this model as an
approach to nonparametric regression on a graph, although with the
simpler case of $\mA = \mI$. He proposes an efficient active-region
based procedure to solve the resulting optimization problem. In the
genomics community, a version of \eqref{eq:r9C33} has been proposed as
the Graph-Guided Fused LASSO problem by \citet*{kim2009multivariate}.
There, the pairwise interactions were estimated from the correlation
structure in the quantitative traits.  Their model is essentially
identical to this one, though with $w_1 = 0$. There, the authors used
a general quadratic solver to tackle the problem and showed good
statistical performance in detecting genetic markers.  The use of
recent results in optimization, in particular proximal operator
methods \citep{nesterov2007gradient}, have been proposed as a way of
optimizing this model \citet*{chen2010graph, chen2012smoothing,
  bach2012structured}.  The resulting algorithms, discovered
independently, are similar to the ones we propose, although we improve
upon them in several ways.

Some recent approaches focus on approximate methods with provable
bounds on the solution.  Nesterov proposed a smooth approximation
approach which can be applied to the graph-guided LASSO in
\citet{nesterov2007gradient}.  More recently, \citet{NIPS2013_4934}
proposed a non-smooth approximation method that decomposes the
summation in \eqref{eq:r9C33}.  

A very similar model to \eqref{eq:r9C33} is proposed in
\citet*{sharma2013consistent}; there, however, the authors include an
additional regularization term penalizing $\absof{u_i + u_j}$.  In
addition, they give several adaptive schemes for choosing the weights
in a way that depends on the correlation structures of the design
matrix. This work is noteworthy in that the authors give a detailed
analysis of the asymptotic convergence and estimation rate of the
model when the regularization weights are chosen adaptively.  Their
optimization method, however, involves a simple quadratic programming
setup, which can severely limit the size of the problems for which
their estimator can be used.

In the context of regularized regression with pairwise dependencies,
our contribution is a theoretical treatment and several algorithms for
a class of models that generalizes \eqref{eq:r9C33}.  Namely, we allow
the $\absof{u_i}$ penalty in the $L_1$ norm to be replaced with an
arbitrary convex piecewise linear function $\xi_i(u_i)$.  In
particular, we examine the estimator
\begin{equation}
  \m u^*_\star = \argmin_{\m u \in \Reals^n} \Lcal\T{\m u, \m y} 
  + \lambda\Tbr{ \sum_i \xi_i(u_i) + \sum_{1 \leq i < j \leq n} w_{ij} \absof{u_i - u_j} }, 
  \label{eq:smUA4}
\end{equation}
which generalizes many of the above models.  While our primary result
is a representation of the theoretical structures underlying this
optimization problem, our work motivates efficient and novel
algorithms for working with these types of structure.  The theory we
develop leads to proofs of the correctness of our algorithms.  In
addition, we are able to present an algorithmic description that gives
the entire regularization path for one version of this problem.

\subsection{Total Variation Models}
\label{sec:b:tv}

Total variation models are effectively an extension of the above
pairwise regularization schemes to the estimation of continuous
functions.  In this context, we wish to estimate a function $u$ in
which the total variation of $u$ -- the integral of the norm of the
gradient -- is controlled.  This model is used heavily in developing
statistical models of images \citep{chambolle2010introduction} and the
estimation of density functions \citep{bardsley2009total}, but arises
in other contexts as well.

When dealing with total variation models, we work theoretically with
general $L_2$-measurable functions with bounded variation.  Our
response becomes a function $f$, and our predictor becomes a function
$u$.  However, in practice, we discretize the problem by working with
the functions $f$ and $u$ only on a discrete lattice of points, as is
common in these problems. 


Let $\Omega \subset \Reals^d$ be a compact region, and let $u$ be an
$L_1$-integrable continuously differentiable function defined on
$\Omega$.  In this context, the total variation of $u$ is given by
\begin{equation}
  \TV(u) = \int_\Omega \lnormof{\grad u}{2} \ud \mu = \int_\Omega \lnormof{(\grad u)(\mx)}{2} \ud\mx.
  \label{eq:tG0cL}
\end{equation}
More general definitions exist when $u$ is not continuously
differentiable \citet{ambrosio2012equivalent,giusti1984minimal}; for
simplicity, we assume this condition.  Let $\Fcal(\Omega)$ represent
the space of $L_1$-integrable continuously differentiable functions of
bounded variation.  Here, bounded variation can be taken as the
condition
\begin{equation}
  \TV(u) < + \infty.
  \label{eq:OrP3C}
\end{equation}
Again, much more general definitions exist, but this suffices for our
purposes.

Formally, then, the total variation problem seeks to find an
estimation function $u^*$ that to a function $f \ST \Reals^d \supset
\Omega \mapsto \Reals$ under a constraint or penalty on the total norm
of the gradient of $u$.
\begin{align}
  u^* &= \argmin_{u \in \Fcal(\Omega)} \sqnorm{u - f} + \lambda \TV(u) \\
  &= \argmin_{u \in \Fcal(\Omega)} \int_\Omega (u(\mx) - f(\mx))^2 \ud\mx + \lambda \TV(u) 
  \label{eq:Mg56R}
\end{align}
This model -- often called the Rudin-Osher-Fatimi (ROF) model after
the authors -- was originally proposed in \citet{rudin1992nonlinear}
as a way of removing noise from images corrupted by Gaussian white
noise; however, it has gained significant attention in a variety of
other areas.

In the image analysis community, it has been used often as a model for
noise removal, with the solution of \eqref{eq:tG0cL} being the
estimation of the noise-free image. In addition, the value of the
total variation term in the optimal solution is often used as a method
of edge detection, as the locations of the non-zero gradients tend to
track regions of significant change in the image.

Beyond image analysis, a number of recent models have used total
variation regularization as a general way to enforce similar regions
of the solution to have a common value.  In this way, it is a
generalization of the behavior of the Fused LASSO in one dimension; it
tends to set regions of the image to constant values.  As such, it is
used in MRI reconstruction, electron tomography
\citep{goris2012electron, gao2010multilevel}, MRI modeling
\citep{michel2011total, keeling2012image}, CT reconstruction
\citep{tian2011low}, and general ill-posed problems
\citep{bardsley2009total}.

\begin{figure}
  \centering
  \begin{subfigure}[b]{0.3\textwidth}
    \centering
    \includegraphics[width=\textwidth]{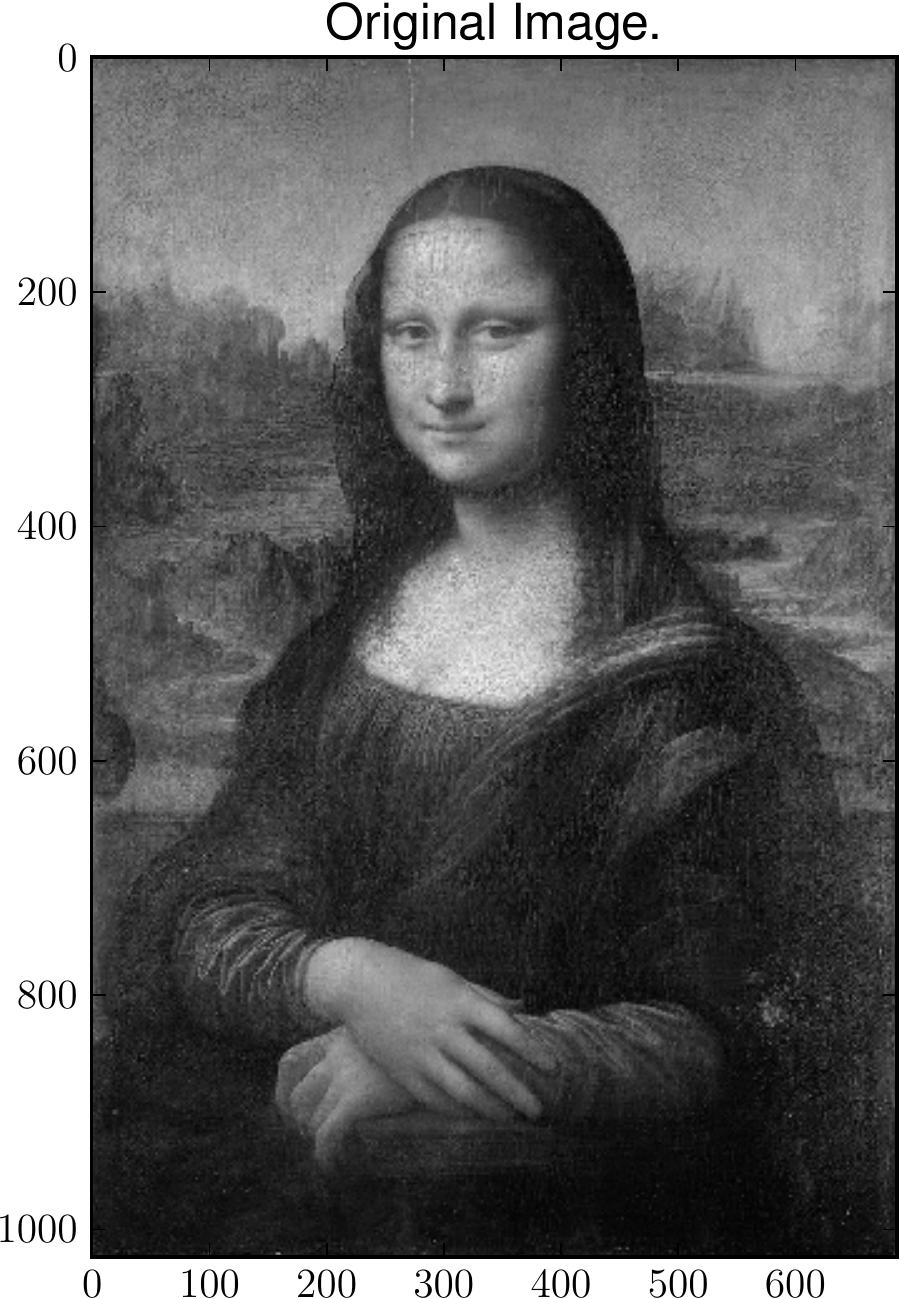}
    \caption{Original Image}
  \end{subfigure}%
  ~ 
  \begin{subfigure}[b]{0.3\textwidth}
    \centering
    \includegraphics[width=\textwidth]{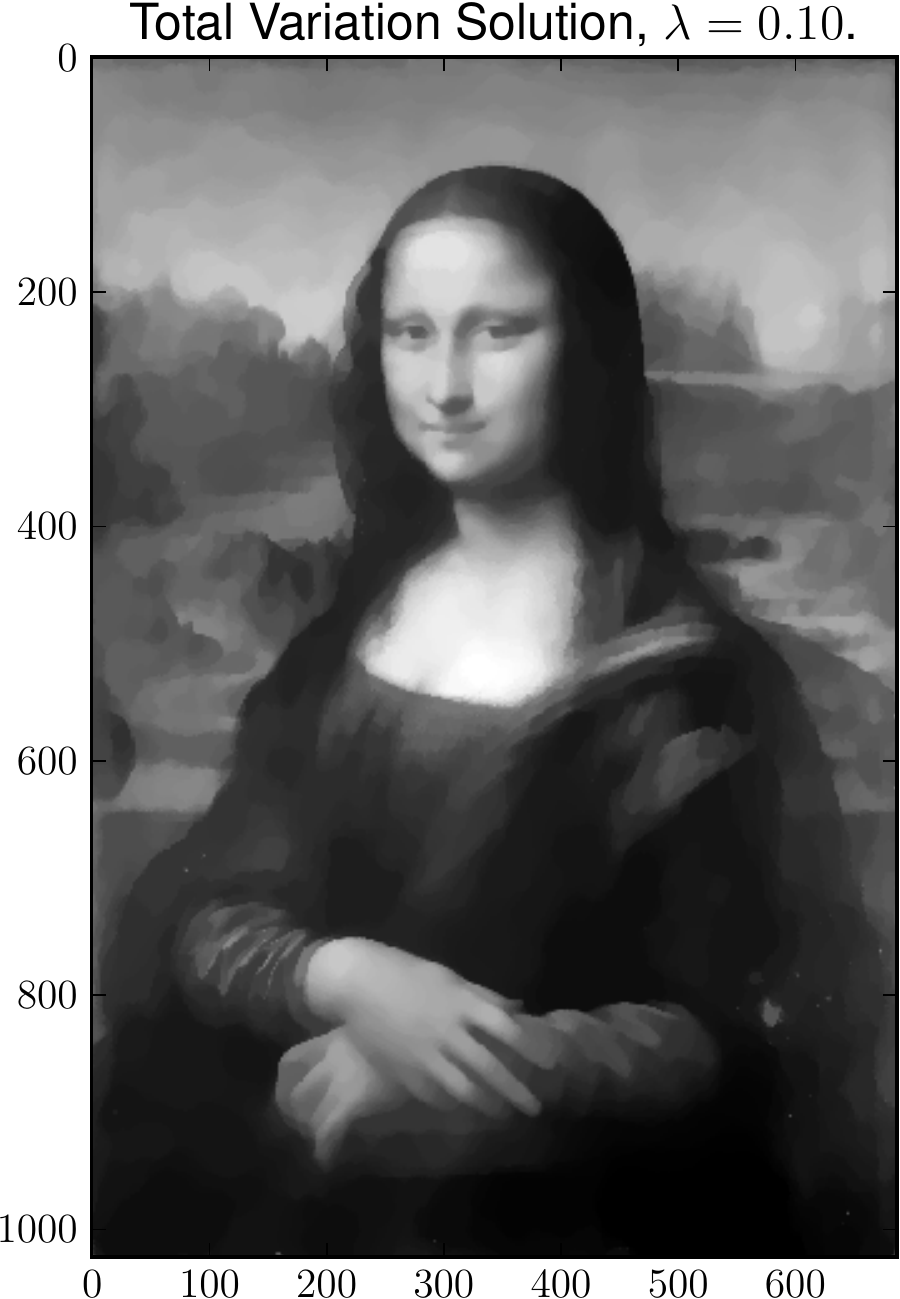}
    \caption{$\TV$ minimized image with $\lambda'=0.1$.}
  \end{subfigure}
  ~ 
  \begin{subfigure}[b]{0.3\textwidth}
    \centering
    \includegraphics[width=\textwidth]{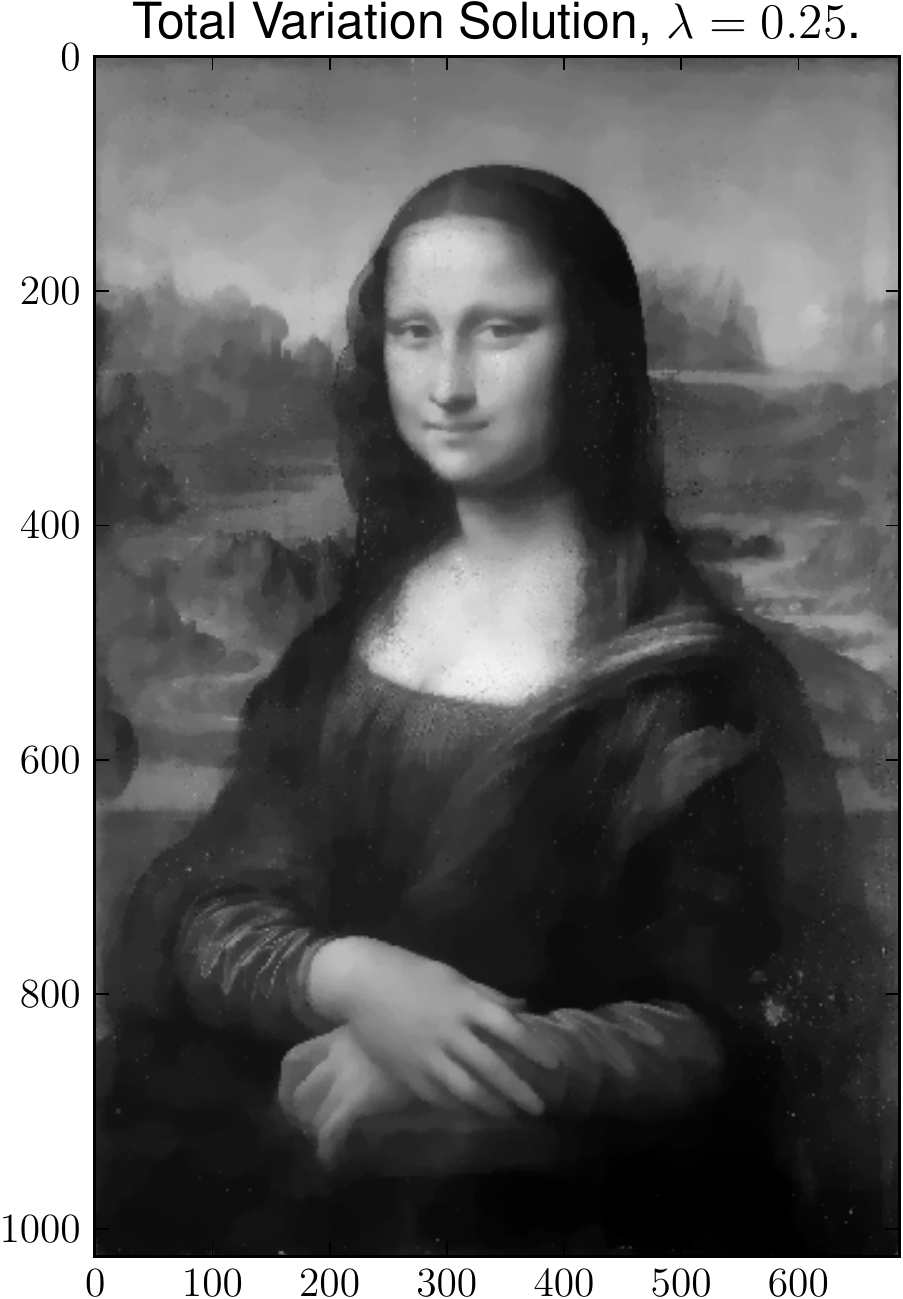}
    \caption{$\TV$ minimized image with $\lambda'=0.25$.}
  \end{subfigure}
  \caption{\small{An example of Total Variation Minimization for noise
      removal on Leonardo da Vinci's Mona Lisa.  Different values of
      the regularization parameter produce different results, with the
      higher value of $\lambda'$ smoothing the image less but removing
      less of the noise.}}
  \label{fig:mona-lisa-simple}
\end{figure}

The total variation regularizer tends to have a smoothing effect
around regions of transition, while also setting similar regions to a
constant value.  As such, it has proven to be quite effective for
removing noise from images and finding the boundaries of sufficiently
distinct regions. The locations where the norm of the gradient
$\TV(u)$ is nonzero correspond to boundaries of the observed process
in which the change is the greatest. A rigorous treatment of the
theoretical aspects surrounding using this as a regularization term
can be found in \citet*{bellettini2002total,ring2000structural} and
\citet*{chan2005aspects}.  A complete and treatment of this topic in
practice can be found in a number of surveys, including
\citet*{darbon2006image, allard2007total, allard2008total,
  allard2009total}; in particular, see \citet*{caselles2011total} and
\citet*{chambolle2010introduction}.  As our purpose in this work is to
present a treatment of the underlying theoretical structures, we refer
the reader to one of the above references for use practical image
analysis.  Still, several points are of statistical interest, which we
discuss now.

Several generalizations of the traditional Rudin-Osher-Fatimi model
have been proposed for different models of the noise and assumptions
about the boundaries of the images. Most of these involve changes to
the loss or log-likelihood term but preserve the total variation term
as the regularizer. In general, we can treat $u^*$ as the estimation
of
\begin{equation}
  u^* = \argmin_{u \in \Fcal(\Omega)} \Lcal(u,f) + \lambda \TV(u)
  \label{eq:wfkV6}
\end{equation}
where $\Lcal$ is a smooth convex loss function.

One option is to use $L_1$ loss for $\Lcal$, namely
\begin{equation}
  \Lcal_1 = \lnormof{f-u}{1},
  \label{eq:zf4i2}
\end{equation}
as a way of making the noise more robust to outliers.  This model is
explored in \citet{bect20041,chan2005aspects} and the survey papers
above.  These perform better under some types of noise and have also
become popular \citep{goldfarb2009parametric}.

Similarly, many models for image denoising -- often when the image
comes from noisy sensing processes in physics, medical imaging, or
astronomy -- involve using a Poisson likelihood for the observations.
Here, we wish to recover a density function where we observe Poisson
counts in a number of cells with rate proportional to the true density
of the underlying process.  Total variation regularization can be
employed to deal with low count rates.  Here, total variation
regularization is used to recover the major structures in the data.
In this case, the optimal loss function is given by
\begin{equation}
  \Lcal(u, f) = \int_\Omega (u(\mx) - f(\mx) \log u(\mx)) \ud\mx,\qqquad u \geq 0.
  \label{eq:Gpwjy}
\end{equation}
Research on this, including optimization techniques, can be found in
\citet{bardsley2009total,bardsley2008efficient,sawatzky2009total}.

Also of interest in the statistics community,
\citet{wunderli2013total} replaces the squared-norm fit penalty in
\eqref{eq:Mg56R} with the quantile regression penalty from
\citet{koenker1978regression}, where loss function is replaced with
\begin{equation}
  \Lcal(u, f) = \absof{f(\mx) - u(\mx)} \times \fOOdbl
  {1 - \beta}{f(\mx) \geq u(\mx)}
  {\beta}{f(\mx) < u(\mx)}.
  \label{eq:Mu9T7}
\end{equation}
for which the author proves the existence, uniqueness, and stability
of the solution. 

One of the variations of total variation minimization is the
Mumford-Shah model.  In this model, used primarily for segmenting
images, regions of discontinuity are handled explicitly
\citet{mumford1989optimal}.  In this model, the energy which is
minimized is
\begin{equation}
  E(u, \Gamma) = \sqnorm{f - u} + \int_{\Omega \backslash \Gamma} \sqnorm{\grad u} \ud \mu + v \normof{\Gamma}
  \label{eq:T6HsV}
\end{equation}
where $\Gamma$ is a collection of curves and $\normof{\Gamma}$ is the
total length of these curves.  Thus the function is allowed to be
discontinuous on $\Gamma$; these are then taken to be the
segmentations of the image.  This model has also attracted a lot of
attention in the vision community as a way of finding boundaries in
the image \citep{chan2000image,pock2009algorithm,el2007graph}.  The
Bayesian model of this is as a type of mixture model in which the mean
function and noise variance differs between regions
\citep{brox2007statistical}.  Unlike general TV regularization,
however, it is typically focused exclusively on finding curves for
image segmentation, rather than on estimating the original image.
While it shares a number of close parallels to our problem,
particularly in the graph cut optimizations used \citep{el2007graph},
our method does not appear to generalize to this problem.


The optimization of the total variation problem has garnered a lot of
attention as well.  An enormous number of approaches have been
proposed for this problem.  These largely fall into the category of
graph based algorithms, active contour methods, or techniques based
around partial differential equations.  These are discussed in more
depth in part 2 of this paper, where we mention them in introducing
our approach to the problem.

\section{Background and Outline of Main Results}
\label{sec:b:outline}

The primary purpose of our work is to present a theory of the
underlying structure connecting the optimization of the above models
over dependent variables and recent results in combinatorial
optimization, particularly submodular function minimization.  Our
theory connects and makes explicit a number of connections between
known results in these fields, many of which we extend in practically
relevant ways.  The insights gained from our theory motivate a family
of novel algorithms for working with these models.  Furthermore, we
are able to give a complete description of the structure of the
regularization path; this result is also completely novel.

One primary practical contribution is the development of methods to
efficiently optimize functions of the following form, which we denote
as \eqref{eq:RB}
\begin{equation}
  \m u^*(\lambda) = \argmin_{\m u \in \Reals^n} \sqnorm{\m u - \m a}
  + \lambda\Tbr{ \sum_i \xi_i(u_i) + \sum_{i,j} w_{ij} \absof{u_i - u_j} }, 
  \label{eq:RB} \tag{$\Re_{\!\text{B}}$}
\end{equation}
with $\ma \in \Reals^n$, $\lambda$ is a non-negative regularization
parameter, $w_{ij} \geq 0$ controls the regularization of dependent
variables, and $\xi_i(u_i)$ is a convex piecewise linear function,
possibly the common $L_1$ penalty $\xi_i(u_i) = \absof{u_i}$.  

We develop a novel algorithm for \eqref{eq:RB} based on network flows
and submodular optimization theory, and completely solve the
regularization path as well over $\lambda > 0$.  The crux of the idea
is to construct a collection of graph-based binary partitioning
problems indexed by a continuous parameter.  With our construction, we
show that the points at which each node exactly gives the solution to
\eqref{eq:RB}.  We develop algorithms for this that scale easily to
millions of variables, and show that the above construction also
unlocks the nature of the regularization path.

These functions arise in two active areas of interest to the
statistics and machine learning communities.  The first is in the
optimization of penalized regression functions where we wish to find
\begin{equation}
  \m u^*(\lambda) = \argmin_{\m u \in \Reals^n} \Lcal\T{\m u, \m y} 
  + \lambda\Tbr{ \sum_i \xi_i(u_i) + \sum_{i,j} w_{ij} \absof{u_i - u_j} }, 
  \label{eq:RL} \tag{$\Re_{\!\text{L}}$}
\end{equation}
where $\m y$ is a collection of observed response and $\Lcal(\m u, \m
y)$ is a smooth, convex log-likelihood term.  Throughout our work, we
denote this problem as \eqref{eq:RL}.  In the general case, we handle
this problem through the use of the proximal gradient methods, a
variant of sub-gradient methods, which we described below.  The inner
routine involves solving \eqref{eq:RB}.

Finally, \eqref{eq:RB} also extends to the continuous variational
estimation problem of Total Variation minimization from section
\ref{sec:b:tv}. where we wish to minimize
\begin{equation}
  u^* = \argmin_{\Fcal(\Omega)} \sqnorm{u - f} + \lambda \TV(u)
  \label{eq:RTV} \tag{$\Re_{\!\text{TV}}$}
\end{equation}
where $\Fcal(\Omega)$ and $\TV(u)$ are defined in \eqref{eq:tG0cL} --
\eqref{eq:Mg56R}. We show in the second part of our work  that this problem can
be solved using the methods developed for \eqref{eq:RB}.  Thus we not
only present an efficient algorithm for \eqref{eq:RTV}, we also
present the first algorithm for finding the full regularization path
of this problem.

\subsection{Outline}
\label{sec:515}

This paper is laid out as followed.  The rest of this section
presents a survey of the background to the current problem, describing
the current state of the research in the relevant areas and laying out
the reasons it is of interest to the statistics and machine learning
communities.  

Section \ref{sec:ch:structure} lays out the basic framework connecting
network flows, submodular function minimization, and the optimization
of \eqref{eq:RB}.  While some of our results were
discovered independently, our theory pulls them together explicitly in
a common framework, and provides more direct proofs of their
correctness.  Additionally, our approach motivates a new algorithm to
exactly solve the resulting optimization problem.

Section \ref{sec:ch:general} presents entirely novel results. We
extend the basic theory of section \ref{sec:ch:structure} using a novel
extension lemma to allow for a general size-biasing measure on the
different components of the input.  The theoretical results are proved
for general submodular functions, extending the state-of-the-art in
this area. In the context of our problem, we use these results to
extend the algorithmic results of section \ref{sec:ch:structure}.  In
particular, this opens a way to handle the $\xi_i(u_i)$ term in
\eqref{eq:RB} above, significantly extending the results of
\citet{mairal2011convex} related to section \ref{sec:ch:structure}.

\subsection{Optimization of Regularized Models}
\label{sec:b:optimization}

Regularized models have gained significant attention in both the
statistics and machine learning communities.  Intuitively, they
provide a way of imposing structure on the problem solution.  This
facilitates the accurate and tractable estimation of high-dimensional
variables or other features, such as change points, that are difficult
to do in a standard regression context.  Furthermore, these methods
have been well studied both from the theoretical perspective --
correct estimation is guaranteed under a number of reasonable
assumptions -- and the optimization perspective -- efficient
algorithms exist to solve them.  Not surprisingly, these approaches
have gained significant popularity in both the statistics and machine
learning communities.  We will describe several relevant examples
below.

In the general context, we consider finding the minimum of
\begin{equation}
  \gamma(\m u) = \Lcal(\m u, \my) + \lambda \Phi(\m u),
  \label{eq:1AZYq}
\end{equation}
where $\Lcal(\m u, \my)$ is a smooth convex Lipshitz loss function,
$\Phi(\m u)$ is a regularization term, and $\lambda \geq 0$ controls
the amount of the regularization.  In statistical models, $\Lcal$ is
typically given by the log-likelihood of the model. The regularization
term $\Phi(\m u)$ is required to be convex but is not required to be
smooth; in practice it is is often a regularization penalty that
enforces some sort of sparseness or structure.  A common
regularization term is simply the $L_1$ norm on $\m u$, i.e.
\begin{equation}
  \Phi_1(\m u)  = \lnormof{\m u}{1}
  \label{eq:1W28K}
\end{equation}
which gives the standard LASSO problem proposed in
\citep{tibshirani1996regression} and described in section
\ref{sec:b:models}. 

A number of algorithms have been proposed for the optimization of this
particular problem \citet{friedman2010regularization,efron2004least}.
Most recently, a method using proximal operators has been proposed
called the Fast Iterative Soft-Threshold Algorithm
\citep{beck2009fast}; this achieves theoretically optimal performance
-- identical to the $\bigOof{\sdfrac{1}{k^2}}$ lower bound of the
convergence rate by iteration of optimization of general smooth,
convex functions \citep{nesterov2005smooth}.  The method is based on
the proximal operators we discuss next, and is described in detail,
along with theoretical guarantees, in \citet{beck2009fast}.  This is
the method that immediately fits with our theoretical results.

\subsubsection{Proximal Operators}
\label{sec:41}

Proximal gradient methods \citet{nesterov2007gradient,beck2009fast}
are a very general approach to optimizing \eqref{eq:1AZYq}.  At a high
level, they can be understood as a natural extension of gradient and
sub-gradient based methods designed to deal efficiently with the
non-smooth component of the optimization.  The motivation behind
proximal methods is the observation that the objective $\gamma(\m u)$
is the composition of two convex functions, with the non-smooth part
isolated in one of the terms.  In this case, the proximal problem
takes the form
\begin{equation}
  \gamma_P(\m u, \mhu) = f(\mhu) + (\m u - \mhu)^T\grad f(\mhu) + \lambda \Phi(\m u) + \frac{L}{2} \sqnorm{\m u - \mhu}
  \label{eq:kOlw7}
\end{equation}
where we abbreviate $f(\m u) = \Lcal(\m u, \m y)$.  $L$ is the
Lipshitz constant of $f$, which we assume is differentiable.  We
rewrite this as
\begin{equation}
  \gamma_P'(\m u, \mhu) 
  = \inv{2} \sqnorm{\m u - \Tbr{\mhu - \inv{L} \grad f(\mhu)} }
  + \frac{\lambda}{L} \Omega(\m u).
  \label{eq:yn6z1}
\end{equation}
The minimizing solution of \eqref{eq:yn6z1} forms the update rule.
Intuitively, at each iteration, proximal methods linearize the
objective function around the current estimate of the solution,
$\mhu$, then update this with the solution of the proximal problem.
As long as \eqref{eq:kOlw7} can be optimized efficiently, the proximal
operator methods give excellent performance.

While this technique has led to a number of algorithmic improvements
for simpler regularizers, it has also opened the door to the practical
use of much more complicated regularization structures.  In
particular, there has been significant interest in more advanced
regularization structures, particularly ones involving dependencies
between the variables.  These types of regularization structures are
the primary focus of our work.

\subsubsection{Related Work}
\label{sec:708}

Finally, of most relevance to our work here, is a series of recent
papers by Francis Bach and others, namely \citet*{bach2012structured},
\citet*{mairal2010network}, and \citet*{mairal2011convex}, that
independently discover some of our results.  In these papers, combined
with the more general work of
\citet*{bach2011optimization,bach2010shaping} and
\citet*{bach2010convex,,bach2011learning}, several of the results we
present here were independently proposed, although from a much
different starting point.  In particular, he proposes using parametric
network flows to solve the proximal operator for $\Phi(\m u) =
\sum_{i,j} w_{ij} \absof{u_i - u_j}$ and that this corresponds to
calculating the minimum norm vector of the associated submodular
function.  These results are discussed in more detail in section
\ref{sec:ch:structure}.

While our results were discovered independently from these, we extend
them in a number of important ways.  First, we propose a new algorithm
to exactly solve the linear parametric flow problem; this is a core
algorithm in these problems.  Second, through the use of the weighting
scheme proposed in section \ref{sec:ch:general}, we develop a way to
include general convex linear functions directly in the optimization
procedure.  This expands the types of regularization terms that can be
handled as part of the proximal operator scheme.

\subsection{Combinatorial Optimization and Network Flows}
\label{sec:b:netflows}

The final piece of background work we wish to present forms the last
pillar on which our theory is built.  Core to the theory and the
algorithms is the simple problem of finding the minimum cost
partitioning of a set of nodes $\Vcal$ in which relationships between
these nodes are defined by a graph connecting them.  Our results are
built on a fundamental connection between a simple combinatorial
problem -- finding a minimum cost cut on a graph -- and the convex
optimization problems described in section \ref{sec:b:optimization}.

A fundamental problem in combinatorial optimization is finding the
minimum cost cut on a directed graph.  This problem is defined by
$(\Vcal, \Ecal, \mc, s, t)$, where $\Vcal$ is a set of nodes and
$\Ecal$ is a set of directed edges connecting two nodes in $\Vcal$.
We consistently use $n$ to denote the number of nodes and, without
loss of generality, assume $\Vcal = \set{1,2,...,n}$, i.e. the nodes
are refered to using the first $n$ counting numbers. Similarly, the
edges are denoted using pairs of vertices; thus $\Ecal \subseteq
\Set{(i,j)}{i,j \in \Vcal}$.  Here, $(i,j)$ denotes an edge in the
graph going from $i$ to $j$.  For an {\it undirected graph}, we simply
assume that $(i,j) \in \Ecal$ if and only if $(j,i) \in \Ecal$, and
that $c_{ij} = c_{ji}$ for all $i,j \in \Vcal$.  $\mc$ associates a
cost with each of the edges in $\Ecal$.  I.e. maps from a pair of
edges to a non-negative cost, i.e. $c_{ij} \in \Reals^+$ for $(i,j)
\in \Ecal$.

The letters $s$ and $t$ here denote specific nodes not in the node set
$\Vcal$.  For reasons that will become clear in a moment, $s$ is
called the {\it source} node and $t$ is called the {\it sink} node.
We treat $s$ and $t$ as valid nodes in the graph and define the cost
associated with an edge from $s$ to $i$ as $c_{si}$; analogously,
$c_{it}$ denotes the cost of an edge from $i$ to the sink $t$.  These
nodes are treated specially in the optimization, however; the {\it
  minimum cut problem} is the problem of finding a cut in the graph
separating $s$ and $t$ such that the total cost of the edges cut is
minimal.  Formally, we wish to find a set $S^* \subseteq \Vcal$
satisfying
\begin{align}
  S^* \in \Argmin_{S \subseteq \Vcal}
  \Tbr{\sum_{\substack{i \in S \\ j \in (\Vcal \backslash S)}} c_{ij}}
  + \Tbr{\sum_{i \in S} c_{it}}
  + \Tbr{\sum_{i \in \Vcal \backslash S} c_{si}} \label{eq:b:mincut}
\end{align}
where $\Argmin$ with a capital $\operatorname{A}$ returns the set of
minimizers, as there may be multiple partitions $S^*$ achieving the
minimum cost.   

The above lays out the basic definitions needed for our work with
graph structures.  This problem is noteworthy, however, as it can be
solved easily by finding the maximal flow on the graph -- one of the
fundamental dualities in combinatorial optimization is the fact that
the cut edges defining the minimum cost partition are the saturated
edges in the maximum flow from $s$ to $t$ in the equivalent network
flow problem.  It is also one of the simplest practical examples of a
submodular function, another critical component of our theory.  We now
describe these concepts.

\subsubsection{Network Flows}
\label{sec:594}

The network flow problem is the problem of pushing as much ``flow'' as
possible through the graph from $s$ to $t$, where the capacity of each
edge is given by the cost mapping $\mc$ above.  The significance of
this problem is the fact that it map directly to finding the minimum
cost partition in a graph \citep{dantzig1955max,
  cormen2001introduction, kolmogorov2004efc}.  The saturating edges of
a maximizing network flow -- those edges limiting any more flow from
being pushed through the graph -- defines the optimal cut in the
minimum cut partitioning.  This result is one of the most practical
results of combinatorial optimization, as many problems map to the
partitioning problem above, and one and simple algorithms exist for
solving network flow problems.

In the network flow problem, we wish to construct a mapping $\mz$ that
represents ``flow'' from $s$ to $t$.  A mapping $\mz$ is a valid flow
if, for all nodes in $\Vcal$, the flow going into the each node is the
same as the flow leaving that node.  Specifically,
\begin{mathDefinition}[(Valid) Flow]
  \label{def:flow} 
  Consider a graph $\Gcal = (\Vcal, \Ecal, \mc, s, t)$ as described
  above.  Then $z_{ij}$, $i, j \in \set{s,t} \cup \Vcal$, is a flow on
  $\Gcal$ if
  \begin{align}
    0 \leq z_{ij} \leq c_{ij} &\Forall i,j \in \Vcal \\
    \sum_{i \in \Vcal \cup \set{s} } z_{ij} = &\sum_{k \in \Vcal \cup
      \set{t} } z_{jk},
  \end{align}
  where for convenience, we assume that $z_{ii} = 0$ and $z_{ij} =
  c_{ij} = 0$ if $(i,j) \notin \Ecal$.  

  Furthermore, we say that an edge $(i,j)$ is saturated if $z_{ij} =
  c_{ij}$, i.e. the flow on that edge cannot be increased.
\end{mathDefinition} %
\nid Since we frequently talk about flows in a more informal sense, we
use the term ``valid flow'' to reference this formal definition.

It is easy to show that the amount of flow leaving $s$ is the same as
the amount of flow leaving $t$, i.e.
\begin{equation}
  \text{Total Flow} = \sum_{i \in \Vcal} z_{si} = \sum_{i \in \Vcal} z_{it}.
  \label{eq:SJTd2}
\end{equation}
this {\it total flow} is what we wish to maximize in the maximum flow
problem, i.e. we wish to find a {\it maximal flow} $\mz^*$ such that
\begin{equation}
  \mz^* \in \Argmin_{\qquad\mz \ST \mz \text{ is a valid flow on $\Gcal$}\qquad} \sum_{i \in \Vcal} z_{si}.
  \label{eq:x4vU7}
\end{equation}
The canonical {\it min-cut-max-flow} theorem \citep{dantzig1955max,
  cormen2001introduction} states that the set of edges saturated by
all possible maximal flows defines a minimum cut in the sense of
\eqref{eq:b:mincut} above.  The immediate practical consequence of
this duality is that we are able to find the minimum cut partitioning
quickly as fast algorithms exist for finding a maximal flow $\mz^*$.
We outline one of these below, but first we generalize the idea of a
{\it valid flow} in two practically relevant ways.

\subsubsection{Preflows and Pseudoflows}
\label{sec:b:pflows}

Two extensions of the idea of a flow $\mz$ on a graph relaxes the
equality in the definition of a flow.  Relaxing the inequality is key
to several algorithms, and forms some interesting connections to the
theory we outline.  The equality constraint is effectively replaced
with an $\excess(\cdot)$ function that gives the excess flow at each
node; if the total amount of flow into and out of a node $i$ is equal,
then $\excess(i) = 0$.  Formally,
\begin{mathDefinition}[Preflow]
  \label{def:preflow}
  Consider a graph $\Gcal = (\Vcal, \Ecal, \mc, s, t)$ as in
  definition \ref{def:flow}. Then $z_{ij}$, $i, j \in \set{s,t} \cup
  \Vcal$, is a flow on $\Gcal$ if
  \begin{align}
    0 \leq z_{ij} \leq c_{ij} &\Forall i,j \in \Vcal \\
    \sum_{i \in \Vcal \cup \set{s} } z_{ij} \geq &\sum_{k \in \Vcal \cup
      \set{t} } z_{jk} \label{eq:preflow:2}.
  \end{align}
\end{mathDefinition} %
\nid With this definition, we define the $\excess(\cdot)$ function as
\begin{equation}
  \excess(i) = \Tbr{\sum_{i \in \Vcal \cup \set{s} } z_{ij}} 
  - \Tbr{\sum_{k \in \Vcal \cup \set{t} } z_{jk}}.
  \label{eq:4ue55}
\end{equation}
It is easy to see that if $\mz$ is a preflow, $\excess(i) \geq 0$ for
all nodes $i$.  Maintaining a valid preflow is one of the key
invariants in the common push-relabel algorithm discussed below.

A {\it Pseudoflow} is the weakest definition of a flow; it relaxes
\eqref{eq:preflow:2} completely, allowing there to be both excesses
and deficits on the nodes of the graph.  However, it does add in an
additional constraint, namely that all edges connected to the source
and sink are saturated.  Formally,
\begin{mathDefinition}[Pseudoflow]
  \label{def:pseudoflow}
  Consider a graph $\Gcal = (\Vcal, \Ecal, \mc, s, t)$ as in
  definition \ref{def:flow}. Then $z_{ij}$, $i, j \in \set{s,t} \cup
  \Vcal$, is a flow on $\Gcal$ if
  \begin{align}
    0 \leq z_{ij} \leq c_{ij} &\Forall i,j \in \Vcal \\
    z_{si} = c_{si} &\Forall i \in \Vcal \\
    z_{it} = c_{it} &\Forall i \in \Vcal
  \end{align}
\end{mathDefinition} %
Note that now the $\excess(i)$ function can be either positive or
negative. A pseudoflow has some interesting geometric and algorithmic
properties, It was described in \citet{hochbaum1998pseudoflow} and an
efficient algorithm for solving network flows based on pseudoflows was
presented in \citet{hochbaum2008pseudoflow}.  We mention it here to
preview our results in section \ref{sec:ch:structure}: the pseudoflow
matches up directly with the base polytope, one of the fundamental
structures in submodular function optimization.

\subsubsection{Network Flow Algorithms}
\label{sec:14}

Finding the minimum cut or the maximum flow on a network is one of the
oldest combinatorial optimization problems, and, as such, it is also
one of the most well studied.  It would be impossible to detail the
numerous algorithms to solve the network flow problem here --
\citet{schrijver2003combinatorial} lists over 25 {\it survey} papers
on different network flow algorithms.  Many of these are tailored for
different types of graphs (sparse vs. dense) and other variations of
the problem.

Perhaps the most well-known algorithm for Network flow analysis is the
push-relabel method.  Variants of it achieve the best known
performance guarantees for a number of problems of interest.  It is
simple to implement; essentially, each node has a specific height
associated with it that represents (informally) the number of edges in
the shortest unsaturated path to the sink.  Each node can have an
excess amount of flow from the source.  At each iteration, a node with
excess either pushes flow along an unsaturated edge to a neighbor with
a lesser height, or increments its height.  Termination occurs when
the source node is incremented to $\sizeof{\Vcal}+1$, where $\sizeof{\Vcal}$
is the number of nodes -- at this point, there is no possible path to
the sink and the maximum cut can be found.  For more information on
this, along with other information on this algorithm, see
\citet{cormen2001introduction} or \citet{schrijver2003combinatorial}.
This algorithm runs at the core of our network flow routines.  This
algorithm is guaranteed to run in polynomial time; in the general
case, it runs in $\bigOof{\sizeof{\Vcal}^2\sizeof{\Ecal}}$ time, where $\Vcal$ is
the set of vertices and $\Ecal$ is the set of edges.  However, it can be
improved to $\bigOof{\sizeof{\Vcal}\sizeof{\Ecal}\log(\sizeof{\Vcal^2} /
  \sizeof{\Ecal})}$ using dynamic tree structures
\citep{schrijver2003combinatorial}.

\subsubsection{The Parametric Flow Problem}
\label{sec:748}

One variation of the regular network flow problem is the {\it
  parametric flow problem} \citep{gallo1989fast}.  The general
parametric flow problem is a simple modification of the maximum flow
problem given above, except that now $c_{si}$ is replaced with a
monotonic non-decreasing function $c_{si}(\beta)$,
\begin{equation}
  \mz^*(\beta) \in \Argmin_{
    \substack{
      \mz \ST \mz \text{ is a valid flow on $\Gcal(\beta)$;} \\
      \qquad\text{Capacity of edge $(s,i)$ given by $c_{si}(\beta)$}\qquad}}
  \sum_{i \in \Vcal} z_{si}.
  \label{eq:x4vU7}
\end{equation}
\citet{gallo1989fast} showed that this problem could be solved for a
fixed sequence $\beta_1 < \beta_2 < \cdots < \beta_m$ in time
proportional to the time of a single run of the network flow problem
by exploiting a nestedness property of the solutions.  Namely, as
$\beta$ increase, the minimum cut of the optimal solution moves
closer on the graph to the source.  More formally, if $\beta_1 <
\beta_2$, then 
\begin{equation}
  S^*(\beta_1) \subseteq S^*(\beta_2), 
  \label{eq:VFzL2}
\end{equation}
where $S^*(\beta)$ is an optimal cut at $\beta$ in the sense of
\eqref{eq:b:mincut}.

If $c_{si}(\beta)$ is a linear function with positive slope, we call
this problem the the {\it linear parametric flow problem}. In sections
\ref{sec:ch:structure} and \ref{sec:ch:general}, we develop an exact algorithm
for this problem that does not require a predetermined sequence of
$\beta$ to solve.  This method is one of the key routines in the
algorithms for general optimization.

\subsubsection{Connections to Statistical Problems}
\label{sec:815}

One of the recent applications of graph partitioning is in finding the
minimum energy state of a binary Markov random field.  In particular,
if we can model the pairwise interactions over binary variables
$x_1,...,x_n$ on the graph using unary and pairwise potential
functions $E^1_i(x_i)$ and $E^2_{ij}(x_i, x_j)$, then the minimum
energy solution can be found using the maximum cut on a specially
formulated graph, provided that $E^2_{ij}(0,0) + E^2_{ij}(1,1) \leq
E^2_{ij}(0,1) + E^2_{ij}(1,0)$ for all $i,j$.  This application of
network flow solvers has had substantial impact in the computer vision
community, where pairwise potential functions satisfying this
condition are quite useful.  

We discuss the details of this application in section
\ref{sec:ch:structure}, where we start with the theory behind these
connections as first building block of our other results.  Ultimately,
we show that network flow algorithms can be used to solve not only
this problem, but the much more general optimization problems
described in section \ref{sec:b:optimization} as well. Through these
connections, we hope to bring the power of network flow solvers into
more common use in the statistical community.

\subsection{Submodular Functions}
\label{sec:b:sfm}

The problem of finding a minimal partition in a graph is a special
case of a much larger class of combinatorial optimization problems
called {\it submodular function minimization.}  Like the problem of
finding a minimal-cost partitioning of a graph given in
\eqref{eq:b:mincut}, this optimization problem involves finding the
minimizing set over a submodular function $f \ST 2^\Vcal \mapsto
\Reals$, where $2^\Vcal$ denotes the collection of all subsets of
a {\it ground set} $\Vcal$.  Given this, $f$ is submodular if, for all sets $S, T
\subseteq \Vcal$, 
\begin{equation}
  f(S) + f(T) \geq f(S \cap T) + f(S \cup T).
  \label{eq:PZiPj}
\end{equation}
This condition is seen as the discrete analogue of convexity.  It is
sufficient to guarantee that a minimizing set $S^*$ can be found in
polynomial time complexity.

An alternative definition of submodularity involves the idea of {\it
  diminishing returns} \citep{fujishige2005submodular}. Specifically
$f$ is submodular if and only if for all $S \subseteq T \subset
\Vcal$, and for all $i \in \Vcal \backslash T$,
\begin{equation}
  f(S \cup \set{i}) -  f(S) \leq f(T \cup \set{i}) -  f(T).
  \label{eq:oXSKK}
\end{equation}
This property is be best illustrated with a simple example.  Let $A_1,
A_2,..., A_n \subseteq \Acal$ be $n$ subsets of a larger set $\Acal$,
and define 
\begin{equation}
  f_{\text{coverage}}(S) = \sizeof{\bigcup_{i \in S} A_i},
  \label{eq:yF7zB}
\end{equation}
so $f_{\text{coverage}}(S)$ measures the coverage of the set $\cup_{i \in
  S} A_i$.  In this context, it is easy to see that
$f_{\text{coverage}}(S)$ satisfies \eqref{eq:oXSKK}.   

$f_{\text{coverage}}(S)$ is an example of a {\it monotone} submodular
function as adding new elements to $S$ is only going to increase the
value of $f_{\text{coverage}}(S)$.  However, many practical examples
do not fall into this category.  In particular, the graph cut problem
given in equation \eqref{eq:b:mincut} above is non-monotone
submodular: 
\begin{equation}
  f_{\text{gc}}(S) = \Tbr{\sum_{\substack{i \in S \\ j \in (\Vcal \backslash S)}} c_{ij}} 
  + \Tbr{\sum_{i \in S} c_{it}}
  + \Tbr{\sum_{i \in \Vcal \backslash S} c_{si}}.
  \label{eq:hLN9X}
\end{equation}
We examine this particular example in more detail in section
\ref{sec:ch:structure}, where we prove $f_{gc}$ is indeed submodular.

Submodular function optimization has gained significant attention
lately in the machine learning community, as many other practical
problems involving the sets can be phrased as submodular optimization
problems.  It has been used for numerous applications in computer
vision, language modeling \citep*{lin2010-submod-sum-nlp,
  hui2012-submodular-shells-summarization}, clustering
\citep*{narasimhan2005-q,narasimhan2007-bcut}, computer vision
\citep{jegelka2011-multilabel-coopcut,jegelka2010-coopcut-image-seg},
and many other domains.  This field is quite active, both in terms of
theory and algorithms, and we contribute some novel results to both
areas in section \ref{sec:ch:general}.

The primary focus of our work has been developing a further connection
between these graph problems and submodular optimization theory.  Our
approach, however, is the reverse of much of the previous work.
Ultimately, we attempted to map the network flow problems back to the
submodular optimization problems.  Surprisingly, this actually opened
the door to several new theoretical results for continuous
optimization, and, in particular, to efficient solutions of the
optimization problems given in section \ref{sec:b:models}.

\subsubsection{Some Formalities}
\label{sec:856}

The theory surrounding submodular optimization, and combinatorial
optimization in general, is quite deep.  Many of the results
underlying submodular optimization require a fairly substantial tour
of the theory of the underlying structures; good coverage of these
results is found in \citep{fujishige2005submodular} and
\citep{schrijver2003combinatorial}.  Most of these results are not
immediately relevant to our work, so we leave them to the interested
reader.  However, several additional results are needed for some of
the proofs we use later.

As with the discussion of problem, we assume that $\Vcal =
\set{1,2,3,...,n}$; our notation intentionally matches that of the
minimum cut problem defined above and is consistent throughout our
work.  We refer to $\Vcal$ as the {\it ground set}. Formally, $f$ can
be restricted to map from a collection of subsets of $2^\Vcal$, which
we refer to consistently as $\Dcal$, with $\Dcal \subseteq 2^\Vcal$.
In the context of submodular functions, $\Dcal$ must be closed under
union and intersection and include $\Vcal$ as an
element\citep{fujishige2005submodular}. In general, and except for
some of our proofs in section \ref{sec:ch:general}, $\Dcal$ can be thought
of as $2^\Vcal$.

\subsubsection{Submodular Function Optimization}
\label{sec:b:sfo}

The first polynomial time algorithm for submodular function
optimization was described in \citep{grotschel1993geometric}; it used
the ellipsoid method from linear programming \citep{chvatal83}.  While
sufficient to prove that the algorithm can be solved in polynomial
time, it was impractical to use on any real problems.  The first {\it
  strongly} polynomial time algorithms -- polynomial in a sense that
does not depend on the values in the function -- were proposed
independently in \citet*{schrijver2000combinatorial} and
\citet*{iwata2001combinatorial}.  The algorithm proposed by Schrijver
runs in $\bigOof{n^8 + \gamma n^7}$, where $\gamma$ refers to the
complexity of evaluating the function.  The latter algorithm is
$\bigOof{\gamma n^7 \log n}$, which may be better or worse depending
on $\gamma$.  The weakly polynomial version of this algorithm runs in
$\bigOof{\gamma n^5 \log M}$, where $M$ is the difference between
maximum and minimum function values. Research in this area, however,
is ongoing -- a strongly polynomial algorithm that runs in
$\bigOof{n^6 + n^5 \gamma}$ has been proposed by
\citet*{orlin2009faster}.

In practice, the minimum norm algorithm -- also called the
Fujishige-Wolfe Algorithm -- is generally much faster, although it
does not have a theoretical upper bound on the running time
\citep{fujishige2005submodular}.  We discuss this algorithm in detail,
as it forms the basis of our work.  However, there are clear cases
that occur in practice where this algorithm does not seem to improve
upon the more complicated deterministic ones -- in
\citet*{jegelka2011-fast-approx-sfm}, a running time of $\bigOof{n^7}$
was reported.  Thus the quest for practical algorithms for this
problem continues; it is an active area of research.

Additionally, several other methods for practical optimization have
been proposed for general submodular optimization or for special cases
that are common in practice.  \citet*{stobbe2010efficient} proposed an
efficient method for submodular functions that can be represented as a
decomposable sum of smaller submodular functions given by
$g_i(\sizeof{U_i \cap S})$, where $g_i$ is convex.  In this particular
case, the function can be mapped to a form that permits the use of
nice numerical optimization techniques; however, many submodular
functions cannot be minimized using this technique, and it can also be
slow \citep{jegelka2011-fast-approx-sfm}.  Along with analyzing the
deficiencies of existing methods, \citet*{jegelka2011-fast-approx-sfm}
propose a powerful approach that relies on approximating the
submodular functions with a sequence of graphs that permit efficient
optimization.  This method is quite efficient on a number of practical
problems, likely because the graph naturally approximates the
underlying structure present in many real-world problems.  

Most recently, in \citet*{iyer2013-fast-submodular-semigradient},
another practical optimization method is proposed; at its core is a
framework for both submodular minimization and maximization based on a
notion of discrete sub-gradients and super-gradients of the function.
While not having a theoretical upper bound itself, it performs
efficiently in practice. This method is also noteworthy in that it can
constrain the solution space in which other exact solvers operate,
providing substantial speedups.

\subsubsection{Geometrical Structures and the Minimum Norm Algorithm}
\label{sec:114}

A number of geometrical structures underpin the theory of submodular
optimization.  In particular, an associated {\it polymatroid} is
defined as the set of points in $\sizeof{\Vcal}$-dimensional Euclidean
space with sums of sets of the dimensions constrained by the function
value of the associated set.  

For notational convenience, for a vector $\mx \in \Reals^n$ and set $S
\subseteq \Vcal$, define
\begin{equation}
  \mx(S) = \sum_{i \in S} x_i.
  \label{eq:daodX}
\end{equation}
In this way, $\mx(S)$ forms a type of unnormalized set measure. 

Now the polymatroid associated with $f$ is defined as
\begin{equation}
  P(f) = \Set{\mx \in \Reals^{\sizeof{\Vcal}}}{\mx(S) \leq f(S) \Forall S \in \Dcal},
  \label{eq:yWo0K}
\end{equation}
$P(f)$ is fundamental to the theory behind submodular function
optimization.  

In our theory, we work primarily with the {\it base} of the
polymatroid $P(f)$, denoted by $B(f)$.  This is the extreme
$(\sizeof{\Vcal} - 1)$-dimensional face of $P(f)$; it is defined as
\begin{equation}
  B(f) = \Set{\mx \in \Reals^{\sizeof{\Vcal}}}{\mx \in P(f), \mx(\Vcal) = f(\Vcal)}.
  \label{eq:PyF4M}
\end{equation}
In the case where the domain $\Dcal = 2^\Vcal$, $B(f)$ is a linear,
convex compact set.  Thus it is often refered to as the {\it Base
  Polytope} of $f$ as it is compact.

The base $B(f)$ is particularly important for our work, as one of the
key results from submodular function optimization is the {\it minimum
  norm algorithm}, which states effectively that sets $S^*$ minimizing
$f$ are given by the sign of the point in $B(f)$ closest to the
origin.  This surprising result, while simple to state, takes a fair
amount of deep theory to prove for general $f$; we state it here:
\begin{mathTheorem}[\citep{fujishige2005submodular}, Lemma 7.4.]
  \label{th:minnorm}
  For submodular function $f$ defined on $2^\Vcal$, let
  \begin{equation}
    \my^* = \argmin_{\my \in B(f)} \lnorm{\my}{2}
    \label{eq:438}
  \end{equation}
  and let $S_1^* = \Set{i}{y_i^* < 0}$ and $S_2^* = \Set{i}{y_i^* \leq
    0}$.  Then both $S_1^*$ and $S_2^*$ minimize $f(S)$ over all
  subsets $S \subseteq \Vcal$.  Furthermore, for all $S^\dagger
  \subseteq \Vcal$ such that $f(S^\dagger) = \min_{S \in \Vcal} f(S)$,
  $S_1^* \subseteq S^\dagger \subseteq S_2^*$.
\end{mathTheorem} %

A central aspect of our theory relies on this result.  In general, it
as one way of working the geometric structure of the problem.  It
turns out that in the case of graph partitioning, $B(f)$ takes on a
particularly nice form.  This allows us to very quickly solve the
minimum norm problem here. We show the resulting minimum norm vector
also gives us the full solution path over weighting by the cardinality
of the minimizing set.  While this basic result was independently
discovered in \citet*{mairal2011convex}, our approach opens several
doors for theoretical and algorithmic improvements, which we outline
in the subsequent sections.

\section{The Combinatorial Structure of Dependent Problems}
\label{sec:ch:structure}

The foundational aspect of our work is the result established in this
section, namely an exploration of network flow minimization problems
in terms of their geometric representation on the base polytope of a
corresponding submodular problem.  This representation is not new; it
is explored in some depth as an illustrative example in
\citep{fujishige2005submodular} and connected to the minimum norm
problems in \citet{mairal2011convex}.  Our contribution, however
minor, is based on a simple transformation of the original problem
that yields a particularly intuitive geometric form.  The fruit of
this transformation, however, is a collection of novel results of
theoretic and algorithmic interest; in particular, we are able to
exactly find the optimal $\m u^*$ over the problem 
\begin{equation}
  \m u^*(\lambda) = \argmin_{\m u \in \Reals^n} \sqnorm{\m u - \m a}
  + \lambda{\sum_{i, j} w_{ij} \absof{u_i - u_j} }.
  \label{eq:12eoo:ch2}
\end{equation}
Recall that this problem was discussed in depth in section
\ref{sec:b:optimization}; here, $\ma \in \Reals^n$ is given, and
$\lambda \in \Reals^+$ and the weights $w_{ij} \in \Reals^+$ control
the regularization.

In this section, we lay the theoretical foundation of this work,
denoting connections to other related or previously known results.
The contribution at the end is a strongly polynomial algorithm for the
solution of a particular parametric flow problem; this algorithm
follows naturally from this representation.  In the next section, we
extend the theory to naturally allow a convex piecewise-linear
function $\xi_i(u_i)$ to be included in \eqref{eq:12eoo:ch2} to match
\eqref{eq:RB}.  


Before presenting the results for the real-valued optimization of
\eqref{eq:12eoo:ch2}, we must first present a number of theoretical
results using the optimization over sets $\Vcal = \set{1,2,...,n}$.
We begin with the simplest version of this optimization -- finding the
minimizing partition of a graph -- and extend this result to general
submodular functions later.


\subsection{Basic Equivalences}
\label{sec:37}

We proceed by defining and establishing equivalences between three
basic forms of the minimum cut problem on a graph.  Recall from
section \ref{sec:b:netflows} that the minimum cut problem is the
problem of finding a set $S^* \subseteq \Vcal$ satisfying
\begin{align}
  S^* \in \Argmin_{S \subseteq \Vcal} 
  \Tbr{\sum_{\substack{i \in S \\ j \in (\Vcal \backslash S)}} c_{ij}}
  + \Tbr{\sum_{i \in S} c_{it}}
  + \Tbr{\sum_{i \in \Vcal \backslash S} c_{si}} \label{eq:s:mincut}.
\end{align}
As we mentioned earlier, several other important problems can be
reduced to this form; in particular, finding the lowest energy state
of a binary Markov random field, when the pairwise potential functions
satisfy the submodularity condition, is equivalent to this problem.
Our task now is to make this explicit.

We here show equivalences between three versions of the problem.  The
first is $\Pcal_E$, which gives the standard energy minimization
formulation, i.e. finding the MAP estimate of
\begin{equation}
  p(\mx) \propto \expof{\sum_{(i,j) \in \Ecal} E^2_{ij}(x_i, x_j) + \sum_{i \in \Vcal} E^1_i(x_i)}
  , \qquad \mx \in \set{0,1}^n
  \label{eq:9r2p4}
\end{equation}
which is equivalent to finding 
\begin{equation}
  \mx^* \in \Argmin_{\mx \in \set{0,1}^n} 
  \sum_{(i,j) \in \Ecal} E^2_{ij}(x_i, x_j)
  + \sum_{i \in \Vcal} E^1_i(x_i).
  \label{eq:vRrMz}
\end{equation}
$\Pcal_Q$ gives the formulation as a quadratic binary minimization
problem; here, the problem is to find $\mx \in \set{0,1}^n$ that
minimizes $\mx^T \mQ \mx$ for an $n\times n$ matrix $\mQ$.  This
version forms a convenient form that simplifies much of the notation
in the later proofs.  Finally, we show it is equivalent to the classic
network flow formulation, denoted $\Pcal_N$.  This states the original
energy minimization problem as the minimum $st$-cut on a specially
formulated graph structure.  The equivalence of these representations
is well known \citep{kolmogorov2004efc} and widely used, particularly
in computer vision applications.  We here present a different
parametrization of the problem which anticipates the rest of our
results.

The other unique aspect of our problem formulation is the use of a
{\it size biasing term}; specifically, we add a term $\beta
\sizeof{S}$ to the optimization problem, where $\beta \in \Reals$ can
be positive or negative.  This term acts similarly to a regularization
term in how it influences the optimization, but to think of it this
way would lead to confusion as the true purpose of this formulation is
revealed later in this section -- ultimately, we show equivalence
between the values of $\beta$ at which the set membership of a node
flips and the optimal values of the continuous optimization problem of
\eqref{eq:RB}.

%
\begin{mathTheorem}
  \label{th:equivalences}
  Let $G = \Vcal, \Ecal)$ be an undirected graph, where $\Vcal$ is the
  set of vertices (assume $\Vcal = \set{1, ..., n}$) and $\Ecal
  \subseteq \Set{(i,j)}{i,j \in \Vcal}$ is the set of edges.  Without
  loss of generality, assume that $i < j \Forall (i,j) \in \Ecal$.
  Define $\Scal_E^*(\beta)$, $\Scal_Q^*(\beta)$, and $\Scal_N^*(\beta)$ as the
  sets of optimizing solutions to the following three problems,
  respectively:
  \begin{description}
  \item[Energy Minimization: $\Pcal_E(\beta)$.] Given an energy
    function $\mE_1 = \T{E_i(x_i) \ST i \in \Vcal}$ defined for each
    vertex $i \in \Vcal$ and a pairwise energy function $\mE_2 =
    \T{E_{i,j}(x_i, x_j) \ST (i,j) \in \Ecal}$ defined for each edge
    $(i,j) \in \Ecal$, with $E_{ij}(0,0) + E_{ij}(1,1) \leq
    E_{ij}(0,1) + E_{ij}(1,0)$, let
    \begin{equation}
      \mX^*(\beta) = \Argmin_{\mx \in \set{0,1}^n} \sum_{i \in \Vcal} \T{E_i(x_i) - \beta x_i}
    + \sum_{\substack{(i,j) \in \Ecal \\ i < j}} E_{ij}(x_i, x_j)
    \label{eq:53}
    \end{equation}
    and let 
    \begin{equation}
      \Scal_E^*(\beta) = \Tcbrss{ \Set{i}{x^*_i = 1} \ST \mx^* \in \mX^*(\beta)}.
      \label{eq:fLilh}
    \end{equation}
    
  \item[Quadratic Binary Formulation: $\Pcal_Q(\beta)$.]  Given
    $\mE_1$ and $\mE_2$ as in $\Pcal_E(\beta)$, define the $n\times n$
    matrix $\mQ = [q_{ij}]$ as:
    \begin{align}
      q_{ij} &= \fOO{E_{ij}(1,1) + E_{ij}(0,0) - E_{ij}(0,1) - E_{ij}(1,0)}{i < j} 
      {0}{\text{otherwise}}
      \\
      q_{ii} &= (E_i(1) - E_i(0)) 
      + \!\!\!\! \sum_{i' < i \ST (i', i) \in \Ecal} \!\!\!\! (E_{i',i}(0,1) - E_{i',i}(0,0)) 
      + \!\!\!\! \sum_{j > i \ST (i, j) \in \Ecal} \!\!\!\! (E_{ij}(1,0) - E_{ij}(0,0)).
    \end{align}
    Suppose $q_{ij} \leq 0$ for $i \neq j$, and let
    \begin{equation}
      \mX^* = \Argmin_{\mx \in \set{0,1}^n} \mx^T (\mQ - \beta\mI)\,\mx, 
      \label{eq:jKiL}
    \end{equation}
    and let 
    \begin{equation}
      \Scal_Q^*(\beta) = \Tcbrss{ \Set{i}{x^*_i = 1} \ST \mx^* \in \mX^*(\beta)}.
      \label{eq:U2al8}
    \end{equation}
  \item[Minimum Cut Formulation: $\Pcal_N(\beta)$.] Let $\Gcal' =
    (\Vcal', \Ecal')$ be an augmented undirected graph with $\Vcal' =
    \Vcal \cup \set{s,t}$, where $s$ and $t$ represent source and sink
    vertices, respectively, and $\Ecal' = \Ecal \cup \Set{(s,i)}{i \in
      \Vcal} \cup \Set{(i,t)}{i \in \Vcal} \cup \Set{(j,i)}{(i,j) \in
      \Ecal}$.  Define capacities $c_{ij}$, $(i,j) \in \Ecal$ on the
    edges as:
    \begin{align}
      c_{si} = c_{si}(\beta) &= \Tbr{a_i(\beta)}^+ &
      c_{ji} = c_{ij} &= -\frac{q_{ij}}{2}, \; i < j &
      c_{jt} = c_{jt}(\beta) &= \Tbr{a_j(\beta)}^-
    \end{align}
    where
    \begin{equation}
      a_i(\beta) = \inv{2} \sum_{i' : i' < i} q_{i',i} + (q_{ii} - \beta) + \inv{2} \sum_{j\ST i e< j} q_{ij}.
      \label{eq:85}
    \end{equation}
    Then the set of minimum cut solutions $\Scal_N^*(\beta)$ is given by
    \begin{equation}
      \Scal_N^*(\beta) = \Argmin_{\substack{S \subset \Vcal' \\ s \in S, \, t \in \Vcal' \backslash S \;\;}} 
      \sum_{(i,j) \in \delta(S, \Vcal' \backslash S) } c_{ij}(\beta).
      \label{eq:SPUJ}
    \end{equation}
  \end{description}
  Then $\Pcal_E(\beta)$, $\Pcal_Q(\beta)$, and $\Pcal_N(\beta)$ are
  equivalent in the sense that any minimizer of one problem is also a
  minimizer of the others; specifically,
  \begin{equation}
    \Scal_E^*(\beta) = \Scal_Q^*(\beta) = \Scal_N^*(\beta) 
    \label{eq:5hNiD}
  \end{equation}
\end{mathTheorem} %
\begin{proof}
  A reformulation of known results \citep{kolmogorov2004efc}, but
  given in section \ref{sec:str:proofs} for convenience.
\end{proof}
The primary consequence of this theorem is that solving the energy
minimization problem can be done efficiently and exactly due to
several types of excellent network flow solvers that make solving
problems with millions of nodes routine \citep{boykov2004experimental,
  cormen2001introduction,schrijver2003combinatorial}.  Because of
this, numerous applications for graphcuts have emerged in recent years
for computer vision and machine learning.  Our purpose, in part, is to
expand the types of problems that can be handled with network flow
solvers, and problems of interest in statistics in particular.

In our work, we alternate frequently between the above
representations. For construction, the energy minimization problem is
nicely behaved.  In the theory, we typically find it easiest to work
with the quadratic binary problem formulation due to the algebraic
simplicity of working in that form.  Again, however, each of these is
equivalent; when it does not matter which form we use, we refer to the
problem an solution set as $\Scal^*(\beta)$ and $\Pcal(\beta)$
respectively.

\subsubsection{Connections to Arbitrary Network Flow Problems}
\label{sec:226}

While theorem \ref{th:equivalences} lays out the equivalence between
$\Pcal_E(\beta)$ and $\Pcal_Q(\beta)$ and a specific form of network
flow problem $\Pcal_N(\beta)$, for completeness we show that any
network flow problem can be translated into the form $\Pcal_N(\beta)$
and thus $\Pcal_E(\beta)$ and $\Pcal_Q(\beta)$.  The two
distinguishing aspects of $\Pcal_N(\beta)$ are the facts that each
node is connected to either the source or the sink, and that all the
edges are symmetric, i.e. $c_{ij} = c_{ji}$ for all $i \neq j$.  It is
thus sufficient to show that an arbitrary flow problem can be
translated to this form.  For conciseness, assume that $\beta = 0$; the
results can be adapted for other $\beta$ easily. 

\begin{mathTheorem}[]
  \label{th:network-flow-equivalences}
  Any minimum cut problem on an arbitrary, possibly directed graph can
  be formulated as a quadratic binary problem of the form
  $\Pcal_Q(\beta = 0)$ as follows:

  \begin{enumerate}[label={\bf \arabic*}., ref=(\arabic*),nolistsep]
  \item \label{th:nfe:1} For all edges $(i,j)$ such that $c_{ij} > c_{ji}$, add a path
    $s \rightarrow j \rightarrow i \rightarrow t$ with capacity
    $c_{ij} - c_{ji}$.  That edge is now undirected in the sense that
    both directions have the same capacity, and the edges in the
    minimum cut are unchanged, as these paths will simply be eliminated
    by flow along that path.
  \item Set $q_{ij} = c_{ij}$ for $i < j$ and $q_{ij} = 0$ for $i>j$.
  \item Given $q_{ij}$, set $q_{ii} = (c_{si} - c_{it}) - 2\Tbr{\sum_{i' : i' < i} q_{i',i} +  \sum_{j\ST i < j} q_{ij}}.$
  \end{enumerate}

  \nid Using these steps, any minimum cut problem can be translated to
  $\Pcal_Q$ in the sense that the set of minimizing solutions is identical.
\end{mathTheorem} %
\begin{proof}
  Network flows are additive in the sense that increasing or
  decreasing the capacity of each edge in any path from $s$ to $t$ by
  a constant amount does not change the set of minimizing solutions of
  the resulting problem, even if new edges are added
  \citep{cormen2001introduction, schrijver2003combinatorial}.  Thus
  step \ref{th:nfe:1} is valid.  The rest follows from simple algebra.
\end{proof}




\subsection{Network Flows and Submodular Optimization}
\label{sec:submodular-optimization}

Recall from section \ref{sec:b:sfm} that the minimum cut problem is a
subclass of the more general class of submodular optimization
problems.  In the context of $\Pcal_Q(\beta)$, it is easy to state a
direct proof of this fact, additionally showing that here the
submodularity of the pairwise terms is also necessary for general
submodularity.
\begin{mathTheorem}
  \label{th:is-submodular}
  The minimization problem $\Pcal_Q(\beta)$ can be expressed as
  minimization of a function $f_\beta \ST 2^\Vcal \mapsto \Reals$,
  with
  \begin{equation}
    f_\beta(S) = \sum_{\substack{i<j \\ i, j \in S}} q_{ij}  + \sum_{i \in S} (q_{ii} - \beta).
    \label{eq:Md7l}
  \end{equation}
  Then $f_\beta$ is submodular if and only if $q_{ij} \leq 0 \Forall
  i, j \in S, i < j$.
\end{mathTheorem} %
\begin{proof}
  One immediate proof of the {\it if} part follows from the fact that
  $\Pcal_Q(\beta)$ can be expressed as the sum of submodular pairwise
  potential terms, and the sum of pairwise submodular functions is
  also submodular \citep{fujishige2005submodular}.  The direct proof,
  including both directions, is a simple reformulation of known
  results see \citep{kolmogorov2004efc}, but given in section
  \ref{sec:str:proofs} on page \pageref{prf:equivalence} for
  convenience.
\end{proof}


\subsubsection{Geometric Structures}
\label{sec:381}





We are now ready to present the theory that explicitly describes the
geometry of $\Pcal_Q(\beta)$, which extends to both $\Pcal_E(\beta)$
and $\Pcal_N(\beta)$, in the context of submodular function
optimization.  This theory is not new; several abstract aspects of it
have been thoroughly explored by \citet{schrijver2003combinatorial}
and \citet{fujishige2005submodular}. In our case, however, the exact
form of the problem presented in theorem \ref{th:equivalences} was
carefully chosen to yield nice properties when this connection is made
explicit.  From these, a number of desirable properties follow
immediately.


The rest of this section is arranged as follows. First, we show that
the base polytope $B(f_\beta)$ is a reduction of all pseudoflows (see
definition \ref{def:pseudoflow}) on the form of the cut problem
$\Pcal_N(\beta)$ from theorem \ref{th:equivalences}.  $B(f_\beta)$,
described in section \ref{sec:b:sfm}, has special characteristics for
our purposes, as the minimum norm algorithm described in section
\ref{sec:b:sfm} provides a convenient theoretical tool to examine the
structure of $\Pcal_Q(\beta)$.  Our key result is to show that the
minimum norm vector -- the $L_2$-projection of the origin onto
$B(f_\beta)$ -- depends on $\beta$ only through a simple, constant
shift.  Thus this vector allows us to immediately compute the minimum
cut directly for any $\beta$, and we are thus able to compute the
minimum cut solution as well for any $\beta$ as well.

\begin{mathTheorem}[Structure of $B(f_\beta)$]
  \label{th:polymatroid}
  Let $f_\beta$ be defined in theorem \ref{th:is-submodular}
  \eqref{eq:Md7l}, and let
  \begin{equation}
    \Acal = \Set{\malpha \in \Mcal_{n \times n}}{
      \fOO
      {\absof{\alpha_{ij}} \leq \absof{q_{ij}}}{i < j} 
      {\alpha_{ij} = 0}{\text{otherwise}}}.
    \label{eq:Acal}
  \end{equation}
  Let $r_i(\malpha)$, $i = 1,...,n$, be defined as follows:
  \begin{equation}
    r_i(\malpha) = q_{ii}
    + \inv{2} \sum_{i' < i} \T{q_{i',i} + \alpha_{i'i}}
    + \inv{2} \sum_{j\ST i < j} \T{q_{ij} - \alpha_{ij}},
    \label{eq:mr}
  \end{equation}
  and denote $\mr(\malpha) = (r_1(\malpha), ..., r_n(\malpha)) \in
  \Reals^n$.  Then the base of the polymatroid associated with
  $f_\beta$, $B(f_\beta)$, is given by
  \begin{equation}
    B(f_\beta) = \Set{\mr(\malpha) - \beta}{\malpha \in \Acal},
  \end{equation}
  and the full polymatroid polytope is given by 
  \begin{equation}
    P(f_\beta) = \Set{\my}{y_i \leq y'_i \Forall i \text{ for some } \my' \in B(f_\beta)}.
    \label{eq:owVE}
  \end{equation}
\end{mathTheorem}
\begin{proof} Proceeds with straightforward albeit tedious algebra.
  Proved in section \ref{sec:str:proofs} on page
  \pageref{sec:str:proofs}. 

\end{proof}

In light of this, the minimum norm problem on the graph structure is
as follows. The immediate corollary to the min-norm theorem is that
$\malpha^*(\beta)$ yields the optimum cut $S^*(\beta)$ of the
corresponding network flow problem, $\Pcal_N(\beta)$:
\begin{mathTheorem}[Minimum Norm Formulation of $\Pcal_N(\beta)$]
  \label{th:mngc-opt}
  Let $\Acal$ and $\mr(\malpha)$ be given by equations \eqref{eq:Acal}
  and \eqref{eq:mr}, respectively.  Then the min-norm problem
  $\Pcal_N(\beta)$ associated with $\Pcal_Q(\beta)$ is defined by
  \begin{equation}
    \malpha^*(\beta) = \argmin_{\malpha \in \Acal} \lnorm{\mr(\malpha) - \beta}{2}
    \label{eq:gcmn}
  \end{equation}
  Then any optimal cut $S^*(\beta)$ solving $\Pcal_G(\beta)$, as given
  in theorem \ref{th:equivalences}, satisfies:
  \begin{equation}
    \Set{i \in \Vcal}{r_i(\malpha^*(\beta)) < \beta} 
    \subseteq S^*(\beta) 
    \subseteq  \Set{i \in \Vcal}{r_i(\malpha^*(\beta)) \leq \beta}.
    \label{eq:b2df}
  \end{equation}
\end{mathTheorem} %
\begin{proof}
  Follows immediately from theorems \ref{th:equivalences} and
  \ref{th:is-submodular} characterizing the cut problem as a
  submodular function optimization problem, theorem
  \ref{th:polymatroid} describing the structure of this problem, and
  theorem \ref{th:minnorm} to characterize the solution.
\end{proof}

The optimal $\malpha^*$ in the above formulation has some surprising
consequences that motivate the rest of our results.  In particular, we
show that the optimal $\malpha^*$ in equation \eqref{eq:gcmn} is
independent of $\beta$; this is the key observation that allows us to
find the entire regularization path over $\beta$.  Formally, this
result is given in theorem \ref{th:opt-cut-solution}.  However, we
need other results first.

\subsubsection{Connection to Flows}
\label{sec:s:flows}
The representation in terms of $\malpha$ is significant partly as the
values of $\malpha$ effectively form a pseudoflow in the sense of
\citet{hochbaum2008pseudoflow} (see section \ref{sec:b:pflows}).
Recall that a pseudoflow extends the concept of a flow by allowing all
the nodes to have both excesses and deficits. In addition, a
pseudoflow assumes that all edges from the source to nodes in the
graph are saturated, possibly creating excesses at these nodes, and
all edges connected to the sink are similarly saturated, possibly
creating deficits at these nodes.

\begin{mathTheorem}[]
  \label{th:bf-flow}
  Consider the problem $\Pcal_N(\beta)$.  For any $\alpha_{ij} \in
  \Reals$, with $(i,j) \in \Ecal$, let $\alpha_{ij}$ represent the
  flow on each edge $c_{ij}$, with $\alpha_{ij} > 0$ indicating flow
  from $i$ to $j$, and $\alpha_{ij} < 0$ indicating flow from $j$ to
  $i$.  Then $\malpha \in \Acal$ defines a pseudoflow on the graph
  structure indexed by $\Acal$. Furthermore, $r_i(\malpha) - \beta =
  \operatorname{excess}(i)$ is the (possibly negative) excess at node
  $i$ in the sense of \eqref{eq:4ue55}.
\end{mathTheorem} %
\begin{proof}
  The pseudoflow condition that all edges from the source node and to
  the sink node are saturated is immediately implied by the fact that
  $r_i(\m0) - \beta w_i = (c_{si} - c_{it}) - \beta$.  The flow
  conditions, then follow from the edge capacity being $c_{ij} =
  c_{ji} = -q_{ij}$ and $-\absof{q_{ij}} \leq \alpha_{ij} \leq
  \absof{q_{ij}}$. 
\end{proof}
\begin{mathCorollary}[]
  \label{cor:}
  Every pseudoflow on the graph defined by theorem
  \ref{th:equivalences} maps to a point in the base polytope
  $B(f_\beta)$, and every point in $B(f_\beta)$ is given by at least
  one pseudoflow.
\end{mathCorollary} %
\begin{proof}
  Follows immediately from theorem \ref{th:bf-flow}.
\end{proof}

\subsection{Structure of the Complete Solution}
\label{sec:583}
The theorem above has a number of important consequences detailed in
the next few sections.  The most immediate consequence comes when we
consider the structure of the optimal solution of the minimum norm
algorithm specialized to the network flow problem $\Pcal_N(\beta)$;
this effectively allows us to derive a way of solving $\Pcal_N(\beta)$
for all $\beta$.
\begin{mathLemma}[Optimal solutions to $\Pcal_N(\beta)$]
  \label{lem:pn-opt}
  Then $\malpha^*$ is an optimal solution to $\Pcal_N(\beta)$ if and
  only if for all $i, j$, $i < j$, the following condition holds:
  \begin{equation}
    \fOOOdbl
    {\qquad \alpha^*_{ij} = \absof{q_{ij}}}
    {\qquad \Iff \qquad r_i(\malpha^*) \geq r_j(\malpha^*)}
    {-\absof{q_{ij}} \leq \alpha^*_{ij} \leq \absof{q_{ij}}} 
    {\qquad \Iff \qquad r_i(\malpha^*) = r_j(\malpha^*)}
    {\qquad \alpha^*_{ij} = -\absof{q_{ij}}}
    {\qquad \Iff \qquad r_i(\malpha^*) \leq r_j(\malpha^*)}.
    \label{eq:24xp}
  \end{equation}
  In particular, the optimum value of $\malpha^*$ in this case is
  independent of $\beta$.
\end{mathLemma} %
\begin{proof}
  First, $\Acal$ is convex as each dimension $\alpha_{ij}$ is bounded
  independently.  Thus the objective of $\Pcal_N(\beta)$ is minimizing
  a convex function over a convex domain.  Therefore, it suffices to
  prove that equation (\ref{eq:24xp}) can be satisfied if and only if
  $\malpha^*$ is a local minimum of $\lnorm{\mr(\malpha) -
    \beta}{2}^2$.  As $\lnorm{\mr(\malpha)-\beta}{2}^2$ is
  differentiable w.r.t. $\malpha$, this is equivalent to showing that
  either the gradient is $0$ or $\malpha^*$ is on the boundary of
  $\Acal$ and all coordinate-wise derivatives point outside the domain
  $\Acal$.

  First, define $g_{ij}(\malpha)$ as the gradient of
  $\lnorm{\mr(\malpha) - \beta}{2}^2$ w.r.t. $\alpha_{ij}$:
\begin{align}
  g_{ij}(\malpha) &= \pder{\alpha_{ij}} \lnorm{\mr(\malpha) - \beta}{2}^2 
  = 2 \sum_k (r_k(\malpha) -\beta) \pder{\alpha_{ij}} r_k(\malpha) 
  \label{eq:0pvq}
\end{align}
Now 
\begin{equation}
    \pder{\alpha_{ij}} r_k(\malpha) = 
    \fOOO{1}{k = i}
    {-1}{k = j}
    {0}{\text{otherwise}}.
    \label{eq:TnWk}
  \end{equation}
  Thus, 
  \begin{align}
    g_{ij}(\malpha) 
    &= 2\Tbr{\T{r_i(\malpha) -\beta} - \T{r_j(\malpha) - \beta}} \\
    &= 2\Tbr{r_i(\malpha) - r_j(\malpha)},
  \end{align}
  and the following condition holds for all $\malpha$:
  \begin{equation}
    \Tcbr{
      \begin{array}{ccc}
        {g_{ij}(\malpha) < 0} \quad & \Iff &    
        \quad {r_i(\malpha) > r_j(\malpha) } \\
        {g_{ij}(\malpha) = 0} \quad & \Iff & 
        \quad {r_i(\malpha) = r_j(\malpha) } \\
        {g_{ij}(\malpha) > 0} \quad & \Iff &       
        \quad {r_i(\malpha) < r_j(\malpha) }
      \end{array}
    }.
    \label{eq:CI5L}
  \end{equation}
  Matching these conditions to those in \eqref{eq:24xp} shows that
  $\malpha^*$ as given defines a local, and thus global, optimum of
  $\Pcal_N(\beta)$. In particular, note that this criterion is
  independent of $\beta$, completing the proof.
\end{proof}

\subsubsection{Invariance to $\beta$}
\label{sec:675}

The invariance of the optimal $\malpha^*$ to $\beta$ allows us to
characterize the solution space of optimal cuts as the level sets of
$\mr(\malpha^*)$.  The core result, as well as our algorithm, is based
on this intuition. 

\begin{mathTheorem}[$ $]

  \label{th:opt-cut-solution}

  \begin{enumerate}[label={\bf \Roman*.}, ref=(\Roman*)]
  \item \label{th:opt-cut-solution:1}
    $\malpha^*$ is the optimal solution to \eqref{eq:gcmn} if and only
    if for all $\beta \in \Reals$, all optimal cuts $S^*_\beta \in
    \Scal^*(\beta)$ for $\Pcal(\beta)$ satisfy
  \begin{equation}
    U_1(\beta) 
    = \Set{i \in \Vcal}{r_i(\malpha^*) < \beta} 
    \subseteq S^*_\beta
    \subseteq  \Set{i \in \Vcal}{r_i(\malpha^*) \leq \beta} 
    = U_2(\beta)
    \label{eq:b2dfb}
  \end{equation}
\item \label{th:opt-cut-solution:2} Furthermore, for all $\beta$,
  $U_1(\beta)$ is the unique smallest minimizer of
  $\Pcal(\beta)$ and $U_2(\beta)$ is the unique largest
  minimizer.
\end{enumerate}
\end{mathTheorem} 
\begin{proof}
  Part \ref{th:opt-cut-solution:1} follows as a direct consequence of
  theorem \ref{th:mngc-opt} and the invariance of the minimum norm
  problem to changes in $\beta$ as given in lemma \ref{lem:pn-opt}.  Part
  \ref{th:opt-cut-solution:2} is then an immediate consequence of the
  minimum norm algorithm.
\end{proof}
This theorem is intuitively important as the key values that permit a
connection to the continuous problems are values of $\beta$ at which
the membership of the different nodes change.  This theorem tells us
that these points are given by the values of the minimum norm vector,
here given as $\mr(\malpha^*)$.

\subsubsection{Monotonicity}
\label{sec:725}

One immediate corollary of theorem \ref{th:opt-cut-solution} is a
monotonicity property on the optimal sets, used in the continuous
optimization theory we present later:
\begin{mathCorollary}
  \label{cor:nestedness}
  Let $\beta_1 < \beta_2$.  Then for all $S^*_1 \in
  \Scal^*(\beta_1)$ and $S^*_2 \in \Scal^*(\beta_2)$,
  \begin{equation}
    S^*_1 \subset S^*_2.
    \label{eq:6Dova}
  \end{equation}
\end{mathCorollary} %
\begin{proof}
  Follows immediately from the equivalence of the optimizing sets to
  the level sets of the minimum norm vector given in theorem
  \ref{th:opt-cut-solution}.
\end{proof}

\subsection{Beyond Network Flows}
\label{sec:508}

Theorem \ref{th:opt-cut-solution} above was discovered independently
for the full case of general submodular functions by
\citet{nagano2011size}.  There, the authors similarly showed that the
level sets of the minimum norm algorithm give the solutions for the
$f(S) - \beta \sizeof{S}$ problem.  While the approach those authors
take is different and more involved, we give a shorter, alternative
proof.  We use a simple argument following from the fact that $B(f)$
constrains the minimum norm vector $\my$ to a constant total sum.  The
result is that the constant offset in the minimum norm objective drops
out of the optimization.  More formally:
\begin{mathTheorem}[Invariance of General Submodular Functions to $\beta$]
  \label{th:min-norm-sfm}

  \begin{enumerate}[label={\bf \Roman*.}, ref=(\Roman*)]
  \item \label{th:min-norm-sfm:1}
  Let $f$ be a general submodular function.  Then $\my^*$ is the
  optimal solution to the minimum norm problem
  \begin{equation}
    \my^* = \argmin_{\my \in B(f)} \sqnorm{\my}
    \label{eq:binZW}
  \end{equation}
  if and only if $\forall\, \beta \in \Reals$, the set of optimizing
  solutions 
  \begin{equation}
    S^*(\beta) = \argmin_{S \in \Dcal} f(S) - \beta \sizeof{S}
    \label{eq:GhQ7t}
  \end{equation}
  satisfies
  \begin{equation}
    U_1(\beta) 
    = \Set{i \in \Vcal}{y^* < \beta} 
    \subseteq S^*(\beta) 
    \subseteq  \Set{i \in \Vcal}{y^* \leq \beta} = U_2(\beta). 
    \label{eq:b2dfb}
  \end{equation}
\item \label{th:min-norm-sfm:2} Furthermore, for all $\beta$,
  $U_1(\beta)$ is the unique minimal solution to \eqref{eq:gcmn} and
  $U_2(\beta)$ is the unique maximal solution in $\Scal^*(\beta)$.
\end{enumerate}
\end{mathTheorem} %
\begin{proof}
  Consider the submodular function $f_\beta(S) = f(S) -
  \beta\sizeof{S}$.  It is easy to show that
  \begin{equation}
    B(f_\beta) = \Set{\mx - \beta\m1}{\mx \in B(f)}. 
    \label{eq:igWSU}
  \end{equation}
  Denote by $\my^*(\beta)$ the minimum norm solution for $f_\beta$.
  Then the submodular problem for $\my^*(\beta)$ is given by 
  \begin{align}
    \my^*(\beta) 
    &= \argmin_{\my \in B(f_\beta)} \sqnorm{\my} \label{eq:binZW:1} \\
    &= \Tbr{\argmin_{\mv \in B(f)} \sqnorm{\mv - \beta\m1}} + \beta\m1 \label{eq:binZW:2} \\
    &= \Tbr{\argmin_{\mv \in B(f)} 
      \Tcbr{\T{\sum_{i \in \Vcal} {v_i}^2}
      - 2\beta\T{\sum_{i \in \Vcal} v_i}}
      + \sizeof{\Vcal}\beta^2} + \beta\m1 \label{eq:binZW:3} \\
    &= \Tbr{\argmin_{\mv \in B(f)} 
      \Tcbr{\T{\sum_{i \in \Vcal} {v_i}^2}} 
      - 2\beta f(\Vcal) 
      + \sizeof{\Vcal}\beta^2} + \beta\m1 \label{eq:binZW:4} \\
    &= \Tbr{\argmin_{\mv \in B(f)} \sqnorm{\mv}} + \beta\m1 \label{eq:binZW:5}
  \end{align}
  where steps \eqref{eq:binZW:2}--\eqref{eq:binZW:3} follow by
  definition of $B(f)$, causing the terms dependent on $\beta$ to drop
  out by way as constants under the optimization.  

  From this, we have that the optimal minimum norm vector for
  $f_\beta$ is just the minimum norm vector for $f$ shifted by
  $\beta$, immediately implying part \ref{th:min-norm-sfm:1}.
  Similarly, part \ref{th:min-norm-sfm:2} follows immediately from the
  minimum norm theorem.
\end{proof}

This result is used in several other sections as well, and has a
number of practical implications for size-constrained optimizations
and related problems.  For a full treatment of related implications,
see \citep{nagano2011size}.

\subsection{Exact Algorithm for Constant Parametric Flows}
\label{sec:algo}

\newcommand\BisectReductions{\textsc{BisectReductions}\xspace}
\begin{algorithm}[thbp]
  \SetKw{Step}{step}
  \dontprintsemicolon
  \caption{\textsc{AlphaReduction}}
  \KwIn{Submodular $\mQ$.}
  \KwOut{$\malpha^*$, the minimizer in $\Acal$ of $\lnorm{\mr(\malpha)}{2}^2$.}
  \Avspace
  
  \tcp{\textrm{\it Begin by calling \BisectReductions below 
      on the full set $\Vcal$ to get $\malpha$.}}
  \Return{\BisectReductions($T=\Vcal$, $\malpha = \m0$, $\mQ$)}
  \Avspace
  \Avspace

  \tcp{\textrm{\it $\malpha_{[T]}$ denotes $\malpha$ restricted to edges with both nodes in $T$.}}
  \Avspace 
  \BisectReductions($T$, $\malpha$, $\mQ$)\;
  \Avspace 
  \Indp

  \lAssign{$r_\mu$}{$\mean_{i \in T} r_i(\malpha)$}
  \Assign{$r_{\min}$}{$\min_{i \in T} r_i(\malpha)$}
  \Avspace

  \lIf{$r_\mu = r_{\min}$}{
    \Return{$\malpha_{[T]}$ \hspace{3em} \texttt{//} \textrm{Done; We are on a single level set.}}\;}
  \Avspace

  \Assign{$\Ecal_T$}{$\Set{(i,j)}{i, j \in T}$}
  \Avspace

  \Assign{$S^*_T$}{Minimum cut on $(T, \Ecal_T)$, with
    capacities formed from $(\mQ_{[T]} - \diag(r_\mu))$ by theorem
    \ref{th:equivalences}.}

  \Avspace 

  \tcp{\textrm{\it Fix the flow on edges in the cut by adjusting the
      source/sink capacities of each node, then removing those
      edges. }}

  \Avspace 
  \lFor{$i \in S_T^*,\, j \in T \backslash S_T^*,\, i<j$}{
    \lAssign{$\alpha_{ij}$}{$-q_{ij}$}
    \lAssign{$q_{ii}$}{$q_{ii} - q_{ij}$} \Assign{$q_{ij}$}{$0$}
  }
  \Avspace 
  \lFor{$i \in T \backslash S_T^*,\, j \in S_T^*,\,  i<j$}{
    \lAssign{$\alpha_{ij}$}{$q_{ij}$}
    \lAssign{$\;\;\,q_{jj}$}{$q_{jj} + q_{ij}$} \Assign{$q_{ij}$}{$0$}
  }

  \Avspace
\tcp{\textrm{\it Recursively solve on the two partitions to fix the other $\malpha$'s. }}
  
  \Avspace
  \Assign{$\malpha_{[S_T^*]}$}{\BisectReductions($S_T^*$, $\malpha$, $\mQ$)}
  \Assign{$\malpha_{[T \backslash S_T^*]}$}{\BisectReductions(
    $T \backslash S_T^*$, $\malpha$, $\mQ$)}
  \Avspace

  \Return{$\malpha_{[T]}$}\;
  \label{algo:reduction}
\end{algorithm}

Using the above theory, we now wish to present a viable algorithm to
calculate the reduction, and hence all cuts, for each node.  The idea
is simple and follows immediately from the similarity of the minimum
cut problem $\Pcal_N(\beta)$ to the structure of $B(f_\beta)$ as
detailed in theorem \ref{th:bf-flow}. As each level set of
$\mr(\malpha^*)$ defines an optimum cut in the graph, we can adjust
all of the unary potentials by $r_\mu = \mean_{i \in S}$, chosen to
bisect the reduction values, and solve the resulting cut problem.  By
the max-flow min-cut theorem, all edges crossing the cut are
saturated.  Specifically,
\begin{align}
  \Forall i,j,\,i<j, &\text{ such that } r_i(\malpha) \leq r_\mu < r_j(\malpha), \; \alpha^*_{ij} = -q_{ij}, \\
  \Forall i,j,\,i<j, &\text{ such that } r_i(\malpha) > r_\mu \geq r_j(\malpha), \; \alpha^*_{ij} = q_{ij} 
  \label{eq:WjIM}
\end{align}
As these $\alpha_{ij}$ are optimal in the final solution $\malpha^*$,
they can be fixed by permanently adding their values to the
corresponding $r_i(\malpha)$ and removing them from consideration in
the optimization.  This then bisects the nodes, allowing us to treat
these two subsets separately when solving for the rest of the
bisections.  The validity of this bisection can also be seen by the
optimality of the minimum norm solution as described by theorem
\ref{th:opt-cut-solution}.

Algorithm \ref{algo:reduction} can be summarized as follows.  We first
start by considering the entire set of nodes, setting the working set
$S = \Vcal$.  At each step, we recursively partition the working set
$S$ using a minimum cut as follows:
\begin{enumerate}[label=\arabic*., ref=(\arabic*), nolistsep]
\item If $\mr(\malpha)$ is constant in $S$, then return. We're done. 
\item Otherwise, set up a network flow problem to bisect the nodes and
  find a minimum cut.  Once a minimum cut is found, set all the edges
  in the cut to their saturated values.
\item Repeat on the two resulting subsets of nodes.
\end{enumerate}
Pseudocode for this algorithm is presented in Algorithm
\ref{algo:reduction}.

\begin{mathTheorem}[Correctness of Algorithm \ref{algo:reduction}.]
  \label{th:algo-correction}
  After the termination of Algorithm \ref{algo:reduction}, all values
  of $\malpha$ are set such that $\lnorm{\mr(\malpha)}{2}$ is
  minimized over $\malpha \in \Acal$.
\end{mathTheorem} %
\begin{proof}
  Let $i,j$, $i\neq j$, be any pair of nodes such that $q_{ij} \neq
  0$, and let $\malpha^\dagger$ be the solution returned by Algorithm
  \ref{algo:reduction}. 
  
  First, suppose that $r_i(\malpha^*) = r_j(\malpha^*)$.  Then
  trivially, the optimality criteria of Lemma \ref{lem:pn-opt} is
  satisfied.  

  Next, suppose that $r_i(\malpha) \neq r_j(\malpha)$, and first
  suppose that $r_i(\malpha) > r_j(\malpha)$. Then, by the termination
  condition of the recursion in \BisectReductions, nodes $i$ and $j$
  must have been separated by a valid minimum cut for some
  $r_\mu$. However, as all edges crossing a minimum cut are saturated
  by the max-flow-min-cut theorem, $\alpha_{ij} =
  \absof{q_{ij}}$. Thus condition \eqref{eq:24xp} in Lemma
  \ref{lem:pn-opt} is satisfied.  Similarly, if $r_i(\malpha) <
  r_j(\malpha)$, then $\alpha_{ij} = -\absof{q_{ij}}$, indicating a
  flow of $\absof{q_{ij}}$ from $j$ to $i$; again, this satisfies
  \eqref{eq:24xp}.

  As the above holds for any pairs of nodes $i, j$, the optimality
  criteria of Lemma \ref{lem:pn-opt} is satisfied globally, proving
  the correctness of the algorithm.
\end{proof}

This algorithm is quite efficient in practice, and it forms an core
routine of the total variation minimization algorithm given in the
second part of our work, where we present full experiments and some
comparisons with existing approaches.

\subsection{Extensions to Real-valued Variables}
\label{sec:767}

One of the intriguing consequences of the above theory, and one that
opens new doors to efficiently optimizing several other classes of
functions, comes as the result of being able to map other correlated
data to this framework.  In general, interactions between terms can be
very difficult to work with in practice.  However, the above theory
allows us to exactly find the optimizer of a large class of general
functions.  These functions may not necessarily be smooth.

Our approach connects closely to several recent results discovered
independently by Mairal \citep{mairal2011convex} and Bach
\citep{bach2010convex}, which connect some of these problems to an
older result by Hochbaum \citep{hochbaum1995strongly}.  The last of
these papers effectively establishes an equivalence between a class of
quadratic objective functions and some types of network flow
algorithms, although the equivalence to parametric flows isn't really
explored.  However, this result was used by Bach in
\citet{bach2010convex} to note that the minimum norm problem of the
network flow problem can be solved using classical methods for solving
parametric flow problems \citep{gallo1989fast}, and the implications
of this for structured sparse recovery are explored in
\citet{mairal2011convex}.  The end result, discovered independently,
parallels the theorem we present below, albeit with a different
algorithm.

In contrast, while less general, the algorithm we presented in
\ref{algo:reduction} gives the exact change points immediately, and
the theoretical framework presented surrounding this problem is more
thoroughly explored here.  However, the primary practical improvement
we provide comes in the next chapter when we incorporate the use of
piecewise-linear convex penalty terms as well.

\begin{mathTheorem}
  \label{th:gen-opt-safe}
  Suppose $\gamma \ST \Reals^n \rightarrow \Reals^+$ can be expressed
  as
  \begin{equation}
    \gamma(\m u) = \sqnorm{\m u - \m a} + \lambda \sum_{i,j} w_{ij} \absof{u_i - u_j}
    \label{eq:12jye}
  \end{equation}
  where $\m u \in \Reals^n$ is the variable we wish to optimize over
  and $\ma \in \Reals^n$, $\lambda > 0$, and $w_{ij} \geq 0$ are
  given. Without loss of generality, assume that $i < j$.  Then the
  minimizer
  \begin{equation}
    \m u^* = \argmin_{\m u \in \Reals^n} \gamma(\m u) 
    \label{eq:P0ynY}
  \end{equation}
  can be found exactly using Algorithm \ref{algo:reduction} with
  \begin{align}
    q_{ii} &= a_i \\
    q_{ij} &= \inv{2} w_{ij} \lambda
  \end{align}
  Then $\m u^* = \mr(\malpha^*)$. 
\end{mathTheorem} %
\begin{proof}
  See appendix \ref{sec:str:proofs}, page \pageref{th:gen-opt-safe:proof}.
\end{proof}

\section{Unary Regularizers and Non-Uniform Size Measures}
\label{sec:ch:general}

In many contexts, it is helpful to use regularization terms or
weighting terms on the solution to control the final behavior of the
result.  In the previous section, we discussed the simplest case in
the submodular function context, namely weighting the problem against
the cardinality of the solution set.  We proved this for the case of
graph-based submodular problems and extended that argument to the full
submodular function context.  In this section, we extend this result
to the case where the size biasing term $\beta \sizeof{S}$ term is
replaced with a weighted size biasing term $\beta \mw(S)$.  Here,
\begin{equation}
  \mw(S) = \sum_{i \in S} w_i \geq 0
  \label{eq:m5iyi}
\end{equation}
is the weighted size biasing term on the function.  We then wish to
find 
\begin{equation}
  \Scal^*(\beta) = \Argmin_{S \subseteq \Vcal} f(S) - \beta \mw(S)
  \label{eq:uicV}
\end{equation}
for all $\beta \in \Reals$ and all positive weight measures $\mw$.  

The results of the rest of these papers are entirely novel.  We prove
that the optimal solution to an alternate construction of the minimum
norm theorem yields the entire solution path for all $\beta$. Like the
last section, we find a vector $\mz^*$ on the base $B(f)$ such that
every solution is given by a zero-crossing of $\Tbr{\mz^* - \beta
  \mw}$.  We then show that this allows us to include more detailed
structures in the continuous optimization problem; in particular, we
are able to incorporate the piecewise-linear convex penalty term
$\xi_i(u_i)$ in \eqref{eq:RB} directly into our optimization.

Our result is based around a very simple technique to augment the
original graph such that auxiliary variables ``attract'' parts of the
regularization influence of $\beta$ and transfer it to associated
variables in the original problem.  We then show that it is possible
to translate this result into arbitrary weights by taking several
well-controlled limits.  The end result is an algorithm for solving
\eqref{eq:uicV} for arbitrary positive weights.  

As this result is novel and holds for general submodular functions, we
prove it for the general case first, then extend it to the special
case of the linear parametric flow problem.  Analogously to algorithm
\ref{algo:reduction}, the algorithm we develop here solves this
problem exactly.  The fact that the solution is exact also allows us
to solve \eqref{eq:RB}.

\subsection{Encoding Weights by Augmentation}
\label{sec:87}

The primary tool used for introducing weights into the optimization of
the level sets of the function is to augment the original problem with
additional variables.  When $\beta = 0$, these variables do not
contribute to the solution values of the base set of nodes.  In the
network flow interpretation, they have no connection to the source or
sink -- but they are subject to the influence by $\beta$ in the
resulting solutions.

With the proper construction, it is possible to guarantee that these
augmented nodes always have the same reduction value as the nodes they
are augmenting; this allows us to construct a graph such that these
values then translate back into weights on the $\beta$ terms.  The
primary tool used is the following lemma, which forms the basis of the
rest of our results. 

\begin{mathLemma}
  \label{lem:integer-weight-addition}
  Let $\Vcal = \set{1,2,...,n}$, and let $f(S)$ be a bounded
  submodular function defined on $\Dcal \subseteq 2^\Vcal$.  (Recall
  that $\Dcal$ is closed under intersection and union.)

  Let $\mw \in \set{1,2,...}^{\sizeof{\Vcal}}$ be a vector of positive
  integer weights, and set $W = \sum_i (w_i - 1)$.  Denote $\Vcal_\mw
  = \Vcal \cup \set{n+1, ..., n+W}$.  Fix $M_\beta\in \Reals^+$ and
  set $M >> M_\beta$ sufficiently large. Then,
  \begin{enumerate}[label={\bf \Roman*}., ref=(\Roman*)]
  \item \label{lem:augment:1} For all $\beta \in \Icc{-M_\beta, M_\beta}$,
    \begin{equation}
      \Argmin_{S \in \Dcal} f(S) - \beta \mw(S) 
      = \Set{T^* \cap \Vcal}{T^* \in \Argmin_{T \subseteq \Vcal_\mw \ST T \cap \Vcal \in \Dcal} f_\mw(T) - \beta\sizeof{T}}
      \label{eq:owVE}
    \end{equation}
    where $\Argmin$ returns the set of minimizing sets, $f_\mw(T)$ is
    submodular and given by
    \begin{equation}
      f_{\mw}(T) = f(T \cap \Vcal) + M \sum_{i \in \Vcal} 
      \sum_{j \in K_i}  \Tbr{\indset{i \in T} + \indset{j \in T} - 2 \ind{ \set{i, j} \subseteq T}}, 
      \label{eq:yev2}
    \end{equation}
    and $K_i$ is a block of indices of length $w_i - 1$, given by
    \begin{equation}
      K_i = \set{\T{n + \sum_{k < i} (w_k - 1)}, ...,  \T{n + \sum_{k < i} (w_k - 1)} + (w_i - 1)}.
      \label{eq:nXK6}
    \end{equation}
  \item \label{lem:augment:2} Define
    \begin{equation}
      \Dcal_\mw = \Set{S \cup T}{S \in \Dcal,\; T = \bigcup_{i \in S} K_i}.
      \label{eq:QNe1}
    \end{equation}
    Then $\Dcal_\mw$ is a distributed lattice and \eqref{eq:owVE} can be
    replaced by 
    \begin{equation}
      \Argmin_{S \in \Dcal} f(S) - \beta \mw(S) 
      = \Set{T^* \cap \Vcal}{T^* \in \Argmin_{T \in \Dcal_\mw} f(T \cap \Vcal) - \beta\sizeof{T}}.
      \label{eq:owVE2}
    \end{equation}
  \item \label{lem:augment:3} Furthermore, 
    \begin{equation}
      f_\mw(T) = f(T \cap \Vcal) \text{ for all } T \in \Dcal_\mw. 
      \label{eq:asw2}
    \end{equation}
  \end{enumerate}

\end{mathLemma} %
\begin{proof}
  See appendix \ref{sec:str:proofs}, page \pageref{lem:integer-weight-addition:proof}. 
\end{proof}

The above lemma is noteworthy as it provides a way to theoretically
augment the original problem in a way that alters the original problem
such that the relative influence of the $\beta$ scaling can be
altered.  In particular, in the augmented problem, the size of the
evaluation set $\sizeof{T}$ includes these augmented nodes -- since
they are included deterministically based on the values in the
unaugmented set $\Vcal$, the unaugmented node is effectively counted
multiple times. This allows us to weight the nodes separately. 

In our context, when dealing with graph structures, this corresponds
to adding a collection of single nodes with no connections other than
an effectively infinite capacity edge connecting each to one of the
base nodes.  As this edge ties the nodes together in any cut solution,
the influence of the global weighting parameter $\beta$ on this
auxiliary node is simply transferred to the attached node. The next two
theorems extend this result to the minimum norm vector $\my^*$, and an
approximation lemma extends this to general weight vectors.

\section{Optimization Structure and The Weighted Minimum Norm Problem}
\label{sec:eubo}

The central result of this section is a weighted version of the
minimum norm problem.  Under this construction, the solution to the
original minimum norm problem is the same as problem a
$\beta\sizeof{S}$ weighting term, but with additional nodes.  However,
the level sets of the resulting vector yield the optimal minimizing
sets $f(S) - \beta \mw(S)$ for all values of the parameter $\beta$.

\begin{mathDefinition}[Weighted Minimum Norm Problem]
  \label{def:weighted-mn}
  For a submodular function $f$ defined on $\Dcal \subseteq
  2^{\Vcal}$, and positive weights $\mw \in \Reals^n$, $\mw > 0$, the
  weighted minimum norm problem is given by 
  \begin{equation}
    \mz^* = \argmin_{\mz \in B(f)} \sum_{i \in \Vcal} \frac{z_i^2}{w_i}.
    \label{eq:Aqts}
  \end{equation}
  and we call the solution vector $\mz^*$ the {\it Weighted Minimum
    Norm Vector}.  Furthermore, if the weights $\mw$ are restricted to
  be positive integers, then this is called the {\it Integer Weighted
    Minimum Norm Problem}.
\end{mathDefinition} %

We use the solution to this problem in an analogous way to the use of
the minimum norm vector of section \ref{sec:ch:structure}.  The
theorems in this section show that $\mz^*$ gives the entire solution
path to $f(S) - \beta \mw(S)$ over $\beta$.

\begin{mathTheorem}[Structure]
  \label{th:min-norm-stuff}
  Let $f$ be a submodular function on $\Dcal$, and let $\mw$, $\Vcal_\mw$,
  $f_\mw$,$K_i$, and $\Dcal_\mw$ be as defined in lemma
  \ref{lem:integer-weight-addition} (in particular, the elements of
  $\mw$ are integers).  Let $\kappa_i = K_i \cup \set{i}$, so
  $\mx\T{\kappa_i} = x_i + \sum_{j \in K_i} x_j$. Then
  \begin{enumerate}[label={\bf \Roman*.}, ref=(\Roman*)]
  \item \label{th:min-norm-stuff:1} The polymatroid associated with
    $f_\mw$ is given by
    \begin{equation}
      P(f_\mw) = \Set{\mx \in \Reals^{\sizeof{\Vcal_\mw}}}{\Forall S \in \Dcal, \;\sum_{i \in S} \mx\T{\kappa_i} \leq f(S)}
      \label{eq:mqgT}
    \end{equation}
    and the associated base polymatroid is given by
    \begin{align}
      B(f_\mw) 
      &= \Set{\mx \in \Reals^{\sizeof{\Vcal_\mw}}}{\mx \in P(f_\mw), \;\mx(\Vcal_\mw) = \sum_{i \in \Vcal} \mx\T{\kappa_i} = f(\Vcal)},
      \label{eq:g7Td:1}
    \end{align}
  \item \label{th:min-norm-stuff:2} Let 
    \begin{equation}
      \my^* = \argmin_{\my \in B(f_\mw)} \lnorm{\my}{2}^2.
      \label{eq:dGE3}
    \end{equation}
    Then 
    \begin{equation}
      y_j^* = y_i^* \text{ for all } j \in K_i
      \label{eq:8mWg}
    \end{equation}
  \item \label{th:min-norm-stuff:3} Furthermore, let $\mz^*$ be the
    solution to the integer weighted minimum norm problem, i.e.
    \begin{equation}
      \mz^* = \argmin_{\mz \in B(f)} \sum_i \frac{z^2_i}{w_i}.
      \label{eq:ekbU}
    \end{equation}
    Then 
    \begin{equation}
      y_i^* = \frac{z_i^*}{w_i} \text{ for all } i \in \Vcal.
      \label{eq:CDfN}
    \end{equation}
    Furthermore a vector $\mz^*$ is the optimal solution to
    \eqref{eq:ekbU} if and only $\my^*$, as given by \eqref{eq:CDfN},
    is the optimal solution to \eqref{eq:dGE3}.
  \end{enumerate}
\end{mathTheorem} %
\begin{proof}
  See appendix \ref{sec:str:proofs}, page \pageref{th:min-norm-stuff:proof}.
\end{proof}

The important concept behind this theorem and its corollaries is that
it demonstrates a direct connection between the minimum norm
problem on the augmented problem $f_\mw$ and the original problem $f$.
This connection allows us to build the theory of the weighted problem
directly upon the original theory, essentially using those results.

\subsection{On the Use of General Positive Weights}
\label{sec:491}

The goal of this section is to extend the above results on integer
$\mw$ to all positive real numbers.  This allows us to do a number of
interesting things, particularly in the case of network flow
algorithms.  It also extends the state of the known theory on general
submodular function minimization outside of the cases we are
interested in.  We here state the form of theorem
\ref{th:min-norm-stuff} for general $\mw$, then discuss some of the
implications for the case of network flows and the continuous
optimization problems introduced earlier.  In
particular, this theorem allows us a way to include the piecewise
linear $\xi_i(u_i)$ term in \eqref{eq:RB}.

\begin{mathTheorem}
  \label{th:general-weights}
  Let $f$ be a submodular function defined on $\Dcal \subset
  2^{\Vcal}$, and let $\mw \in \Reals^{\sizeof{\Vcal}}$ be strictly
  positive, finite weights.  Then 
  \begin{enumerate}[label={\bf \Roman*.}, ref=(\Roman*)]
  \item \label{th:general-weights:1} Let $\mz^*$ be the optimal
    solution to the weighted minimum norm problem, i.e.
    \begin{equation}
      \mz^* = \argmin_{\mz \in B(f)} \sum_{i \in \Vcal} \frac{z_i^2}{w_i},
      \label{eq:SIyy}
    \end{equation}
    and, for all $\beta \in \Reals$, let
    \begin{align}
      U_1(\beta) &= \Set{i \in \Vcal}{z^*_i - \beta w_i < 0} \label{eq:33eo:1} \\
      U_2(\beta) &= \Set{i \in \Vcal}{z^*_i - \beta w_i \leq 0} \label{eq:33eo:2}
    \end{align}
    and let 
    \begin{equation}
      \Scal^*(\beta, \mw) = \Argmin_{S \in \Dcal} f(S) - \beta \mw(S).
      \label{eq:N6va}
    \end{equation}
    Then $\mz^*$ is the optimal solution to \eqref{eq:SIyy} if and
    only if, for all $\beta \in \Reals$ and all $S^* \in \Scal^*(\beta, \mw)$,
    \begin{equation}
      U_1(\beta) \subseteq S^* \subseteq U_2(\beta) 
      \label{eq:LQ4r}
    \end{equation}
  \item \label{th:general-weights:2} Furthermore, for all $\beta$,
    $U_1(\beta)$ is the unique minimal solution to \eqref{eq:N6va} and
    $U_2(\beta)$ is the unique maximal solution in $\Scal^*(\beta, \mw)$.
  \end{enumerate}
\end{mathTheorem} %
\begin{proof}
  This result is somewhat involved and quite technical.  We present a
  proof of it, along with supporting lemmas, in section
  \ref{sec:proof-wmn} on page \pageref{sec:proof-wmn}. 
\end{proof}

The above problem allows us to generalize the previous results of
size-penalized submodular optimization to general weighted penalties.
This result may have several significant practical implications;
several of these we explore later in the context of the network flow
analysis results.

An interesting corollary to the above theorem is that the original
formulation of the minimum norm problem is still valid when the norm
being optimized over is reweighted.  It may be that this would open up
an way to remove some of the numerical difficulties often encountered
with the minimum norm problem \citep{jegelka2011-fast-approx-sfm}.
More formally,

\begin{mathCorollary}[Validity of Weighted Minimum Norm.]
  \label{cor:min-norm-connection}
  Let $\mz^*$ be the optimal value of the weighted minimum norm
  problem, with $\mw \in \Reals^{\sizeof{\Vcal}}$, $\mw > 0$.  Let
  $U_1 = \Set{i}{z_i^* < 0}$ and $U_2 = \Set{i}{z_i^* \leq 0}$, Then
  both $U_1$ and $U_2$ minimize $f(U)$ over all subsets $U \in \Dcal$.
  Furthermore, for all $U^\dagger \in \Dcal$ such that $f(U^\dagger) =
  \min_{U \in \Vcal}$, $U_1 \subseteq U^\dagger \subseteq U_2$.  In
  other words, the weighted minimum norm vector $\mz^*$ may be
  substituted for the original minimum norm vector.
\end{mathCorollary} %
\begin{proof}
  Set $\beta = 0$ in theorem \ref{th:general-weights}. 
\end{proof}

\subsubsection{Handling the Case of $w_i = 0$}
\label{sec:709}

One of the challenging aspects here is that we might be interested in
the case of $w_i = 0$.  In theory, this can be easily handled by
simply allowing $w_i$ to be so small that its effect on the problem is
negligibly different from $w_i = 0$; in other words, we can see it as
the limit $w_i \convdown 0$. In practice, this leads to numerical
issues. Thus we propose here a numerically stable method to work with
$w_i = 0$ by investigating the limiting behavior. 

\begin{mathTheorem}
  \label{th:zero-approximation}
  Let $\mw \geq 0$ and define $Q = \Set{i \in \Vcal}{w_i = 0}$.  Then let
  \begin{equation}
    \mz^*_Q = \min_{\mz \in B(f)} \sum_{i \in Q} z_i^2.
    \label{eq:GKmlv}
  \end{equation}
  and let 
  \begin{equation}
    \mz^* = \argmin_{\substack{\mz \in B(f) \\ \sum_{i \in Q} \lnorm{\mz[Q]}{2} = \lnorm{\mz_{Q}^*[Q]}{2}}} \sum_{i \notin Q} \frac{z_i^2}{w_i}.
    \label{eq:liOkR}
  \end{equation}
  where $\mz[Q]$ denotes the vector of elements of $\mz$ in $Q$.  Then, for
  \begin{equation}
    \mz_\eps^* = \argmin_{\mz \in B(f)} \sum_{i \in \Vcal} \frac{z_i^2}{\max\T{\eps, w_i}},
    \label{eq:ghnhu}
  \end{equation}
  we have that 
  \begin{equation}
    \mz_\eps^* \goesto \mz^* \text{ as } \eps \convdown 0
    \label{eq:ksXfF}
  \end{equation}
\end{mathTheorem} %
\begin{proof}
  Consider the form of the optimization problem in \eqref{eq:ghnhu}.
  For $\eps$ sufficiently small, we have that 
  \begin{equation}
    \mz_\eps^* 
    = \argmin_{\mz \in B(f)} 
    \Tbr{\sum_{i \notin Q} \frac{z_i^2}{w_i}} + \eps^{-1} \Tbr{\sum_{i \in Q} z_i^2}
    \label{eq:rm0iD}
  \end{equation}
  As $\eps^{-1} \convup \infty$, the optimum value of $\mz$ is
  constrained to be on the simplex in which $\sum_{i \in Q} z_i^2$ is
  minimal.  This value is given by $C_Q$ above, and this is the
  constraint that is enforced explicitly in \eqref{eq:liOkR}.  Since
  everything is continuous, it is valid to take the limit as $\eps
  \convdown 0$.  Thus the theorem is proved.
\end{proof}

It is outside the current realm of our investigation how to implement
this in the inner workings of the general minimum norm algorithm;
however, we will revisit this issue later when proving the correctness
of the network flow version of the weighted reduction algorithm. 

\section{Network Flow Solutions to and Unary Regularizers}
\label{sec:gw:nf}

We now turn our attention to the specific case of network flows. The
linear parametric flow problem is similar to the flow problem
described earlier, except that now we allow the capacity functions --
analogous to the unary energy terms -- to be a non-decreasing linear
function of the weighting term $\beta$.  Previously, we treated this
global weighting term as having equal influence on all nodes.  This
section considers the case where the influence of $\beta$ has a
different weight on each node. Specifically, we replace $\beta$ with
$\beta w_i$ in theorem \ref{th:equivalences}.  In this case, we are
still able to compute the entire path directly.  This result, while
interesting in its own right, also sets the stage for our later
results for total variation minimization.

To be specific, we extend the problems in theorem
\ref{th:equivalences} as follows:

\begin{description}
\item[Energy Minimization: $\Pcal_E(\beta, \mw)$.] Let $\mE_1$ and
  $\mE_2$ be defined as in theorem \ref{th:equivalences}, and let
  \begin{equation}
    \mX^*(\beta, \mw) = \Argmin_{\mx \in \set{0,1}^n} \sum_{i \in \Vcal} \T{E_i(x_i) - \beta w_i x_i}
    + \sum_{\substack{(i,j) \in \Ecal \\ i < j}} E_{ij}(x_i, x_j)
    \label{eq:53eou}
  \end{equation}
  and let 
  \begin{equation}
      \Scal_E^*(\beta, \mw) = \Tcbrss{ \Set{i}{x^*_i = 1} \ST \mx^* \in \mX^*(\beta, \mw)}.
    \label{eq:YCwVg}
  \end{equation}

\item[Quadratic Binary Formulation: $\Pcal_Q(\beta, \mw)$.]  Given
  $\mQ$ defined as in theorem \ref{th:equivalences}, let
  \begin{equation}
    \mX^*(\beta, \mw) = \Argmin_{\mx \in \set{0,1}^n} \mx^T \T{\mQ - \beta\diag{\mw}}\,\mx, 
    \label{eq:jKi33}
  \end{equation}
  and let
  \begin{equation}
    \Scal_Q^*(\beta, \mw) = \Tcbrss{ \Set{i}{x^*_i = 1} \ST \mx^* \in \mX^*(\beta, \mw)}.
    \label{eq:YCwVg}
  \end{equation}

\item[Minimum Cut Formulation: $\Pcal_C(\beta, \mw)$.] Let the graph
  structure be defined as in theorem \ref{th:equivalences}, and
  define capacities $c_{ij}$, $(i,j) \in \Ecal$ on the edges as:
  \begin{align}
    c_{si} = c_{si}(\beta) &= \Tbr{a_i(\beta, \mw)}^+ &
    c_{ji} = c_{ij} &= -\frac{q_{ij}}{2}, \; i < j &
    c_{jt} = c_{jt}(\beta) &= \Tbr{a_j(\beta,\mw)}^-
  \end{align}
  where
  \begin{equation}
    a_i(\beta, \mw) = \inv{2} \sum_{i' : i' < i} q_{i',i} + (q_{ii} - \beta w_i) + \inv{2} \sum_{j\ST i < j} q_{ij}.
    \label{eq:85}
  \end{equation}
  Then the minimum cut solution $S_C^*(\beta)$ is given by
  \begin{equation}
    S_C^*(\beta) = \Argmin_{\substack{S \subset \Vcal' \\ s \in S, \, t \in \Vcal' \backslash S \;\;}} 
    \sum_{(i,j) \in \delta(S, \Vcal' \backslash S) } c_{ij}(\beta, \mw).
    \label{eq:SPUJ}
  \end{equation}
\end{description}

It is a simple matter to show that the above are equivalent:
\begin{mathTheorem}[]
  \label{th:it-works-really}
  $\Pcal_E^*(\beta, \mw)$, $\Pcal_Q^*(\beta, \mw)$, and $\Pcal_C^*(\beta, \mw)$
  from the above description are equivalent in the sense that any
  minimizer of one problem is also a minimizer of the others, i.e.
  \begin{equation}
    \Scal_E^*(\beta, \mw) = \Scal_Q^*(\beta, \mw) = \Scal_C^*(\beta, \mw).
    \label{eq:bxD0g}
  \end{equation}
\end{mathTheorem} %
\begin{proof}
  Replace $\beta$ with $\beta w_i$ or $\beta \mw$ as appropriate in
  the proof of theorem \ref{th:equivalences}.
\end{proof}

However, it is a much more complicated endeavor to show that this
formulation can be solved exactly in a similar manner to the previous
result.  In the end, we prove the following result: analogously to
before, we find an optimal pseudoflow $\malpha^*$ such that the zero
crossings of $\mr(\malpha^*) - \beta \mw$, with $\mr(\malpha)$ defined
as in \ref{th:polymatroid}, give the level sets of the augmented
problem. The purpose of the current section is to define these
relationships explicitly.

The above result works as well for general network flow solutions as
well.  In this case, we have the following immediate corollary to
theorem \ref{th:general-weights}:

\begin{mathCorollary}[]
  \label{cor:opt-weighted-cut-solution}
  Let $\mw$ be a collection of positive weights.  Then $\malpha^*$ is the
  optimal pseudoflow solution to the weighted minimum norm problem
  \begin{equation}
    \malpha^* = \argmin_{\malpha \in \Acal} \sum_{i \in \Vcal} \frac{r^2(\malpha)}{w_i} 
    \label{eq:uawd}
  \end{equation}
  if and only if
  $\forall\, \beta \in \Reals$, the optimal cut $S^*(\beta, \mw)$ for
  $\Pcal_G$ satisfies
  \begin{equation}
    \Set{i \in \Vcal}{r_i(\malpha^*) < \beta w_i} 
    \subseteq S^*(\beta, \mw) 
    \subseteq  \Set{i \in \Vcal}{r_i(\malpha^*) \leq \beta w_i}.
    \label{eq:b2dfb}
  \end{equation}
\end{mathCorollary} %
\begin{proof}
  Replace $B(f)$ with its representation for network flow problems as
  given in theorem \ref{th:polymatroid}.
\end{proof}

An alternate version of this, which is handy for the proofs, is the
following: 
\begin{mathCorollary}[]
  \label{cor:opt-weighted-cut-solution:alt}
  $\malpha^\dagger$ is optimal for a problem if and only if, for all $\beta
  \in \Reals$, the following condition holds for each ordered pair $i,
  j \in \Vcal$, $i < j$:
  \begin{align}
    \text{if}\qquad r_i(\malpha^\dagger) - \beta w_i \leq 0 < r_i(\malpha^\dagger) - \beta w_j,
    \qquad&\text{then}\qquad \alpha_{ij} = -\absof{q_{ij}} \label{eq:joqoi:1} \\
    \text{if}\qquad r_i(\malpha^\dagger) - \beta w_i > 0 \geq r_i(\malpha^\dagger) - \beta w_j,
    \qquad&\text{then}\qquad \alpha_{ij} = \absof{q_{ij}} \label{eq:joqoi:2}
  \end{align}
\end{mathCorollary} %
\begin{proof}
  Follows immediately from noting that criteria \eqref{eq:b2dfb} in
  corollary \ref{cor:opt-weighted-cut-solution} specifies that if there
  exists a $\beta$ that separates two scaled reductions, then there is
  valid cut in the associated flow problem that separates these nodes.
  This, however, is equivalent to the condition on the $\alpha$ terms
  given by \eqref{eq:joqoi:1} or \eqref{eq:joqoi:2}. 
\end{proof}

\subsubsection{Anchoring Nodes with Differing or Infinite Weights}
\label{sec:915}

While the original interpretation of the weights is to vary the
influence of each node in the regularization path, a natural extension
in the network flow context is to use these weights to fix nodes at
particular reduction values -- informally, to make them less
responsive to the influence of the external flow during the
optimization.  This is done by scaling both the base reduction value
$q_{ii}$ and the $\beta$ term by $w_i$ instead of just $\beta$. In
this context, then, replace $\mr(\malpha)$ with $\mr^\mw(\malpha)$,
defined as
\begin{equation}
  r_i^\mw(\malpha) = w_i q_{ii} + \inv{2} \Tbr{
    \sum_{i' < i} \T{q_{i',i} + \alpha_{i'i}}
    + \sum_{j\ST i < j} \T{q_{ij} - \alpha_{ij}}}
  \label{eq:tifTf}
\end{equation}
and the node membership changes sign at 
\begin{align}
  S^*(\beta) 
  = \Indset{r_i^\mw(\malpha^*) - \beta w_i \leq 0} \\
  = \Indset{\Tbr{q_{ii} - \beta} + \inv{2 w_i} \Tbr{
    \sum_{i' < i} \T{q_{i',i} + \alpha^*_{i'i}}
    + \sum_{j\ST i < j} \T{q_{ij} - \alpha^*_{ij}}}}
  \label{eq:OTWm6}
\end{align}
As a result, it is possible to anchor nodes at a particular value by
setting the corresponding weights to a high value and adjusting the
corresponding source-to-sink weights to match.  In particular, taking
the limit as $w_i \goesto \infty$ in \eqref{eq:OTWm6} causes the
membership to be fixed at the internal reduction value.  In this case,
it effectively fixes these nodes at a pre-determined reduction value,
even though the node still influences other nodes in the flow.  In
particular, this fact is used below to handle the case where we wish
to look at the $L_1$ norm of a reduction value, as we explain later.

\subsubsection{Algorithm}
\label{sec:728}

In the unweighted algorithm, one of the key routines was shifting the
reduction level of all the nodes by the average on that region.  This
allowed us to effectively bisect the problem at that value of $\beta$;
this in turn allowed us to set some of the edges as saturated, fixing
them at their extreme values.  When the resulting solution on a region
is that the flow solution sets all reductions to the average value --
zero in the scaled version -- that region is connected in the final
solution, and we can set it to its common value.

Now that we are putting weights on the nodes, the idea of an
``average'' is redefined.  Now, it is the value of the $\beta_\mu$
such that, if all the reductions values were in the same contiguous
block, $r_i(\malpha) - \beta_\mu w_i$ would equal $0$.  This means
that when $\beta$ is increased or decreased, all of these nodes would
flip at the same time. 

Formally, this $\beta$ value on a set $T$ is given by:
\begin{equation}
  \beta_\mu(T, \malpha) = \frac{\sum_{i \in T} r_i\T{\malpha} w_i}{\sum_{i \in T} w_i}.
  \label{eq:bh6CC}
\end{equation}
This behavior is consistent with the augmentation scheme discussed
earlier -- a node with weight $2$ would count the same as $2$ nodes in
the ``average.''

Now, when solving the resulting flow equation, it is important that
the reduction values are scaled away from the mean such that shifts in
the flow affect the reduction value on the correct scale. 
Since all we care about are the zero-crossings of the nodes, we then
rescale all the starting reduction values away from the mean by
$w_i^{-1}$.  In other words, we end up setting the unary potential
value $\rho_i$ used in calculating the cut to
\begin{equation}
  \rho_i = \inv{w_i} \Tbr{w_i q_{ii} - \beta_\mu(T,\malpha)} = q_{ii} - \inv{w_i}\beta_\mu(T) 
  \label{eq:Uq6VX}
\end{equation}
Note that $\sum_{i \in T} w_i \rho_i = 0$, as we would expect, and
that when used to split the regions, $\beta_\mu$ will always separate
the starting reductions $\rho_i$ into positive and negative
components, ensuring a non-trivial cut solution when this is
incorporated into the algorithm.  As before, the resulting cut
dictates the edges that will be saturated; this is how the solution in
the rest of the problem is tracked.

In the case of $w_i = 0$, these nodes simply do not enter into the
average; In rescaling them away from the mean, we set them to extreme
positive or negative numerical values.  This approach is consistent
with theorem \ref{th:zero-approximation}; it will force the squared
reduction value in these nodes to be minimized.

Thus we can define a weighted version of the bisection routine,
\textsc{WeightedBisectionCut}, to work with weighted nodes:

\begin{algorithm}[htb]
  \SetKw{Step}{step}
  \dontprintsemicolon
  \caption{\textsc{WeightedBisectionCut}}
  \KwIn{Submodular $\mQ$, weights $\mw$, subset of nodes $T$.}
  \KwOut{Bisecting cut $S\subseteq T$ or $\emptyset$ if all nodes in same partition.}

  \Assign{$\mrho$}{$\m0$}

  \If{$\sum_{i\in T} w_i = 0$}{

    \Avspace \tcp{\textrm{In this case, the optimal cut that divides
        the reductions into the positive or negative components is all
        that matters; $w_i = 0$ means this result is true for all
        $\beta$.}}
    
    \lFor{$i \in T$}{ 
      \lAssign{$\rho_i$}{$q_{ii}$}}
  } \Else { 

    \Avspace

    \Assign{$\mu$}{\DS{\frac{\sum_{i \in T} q_{ii} w_i}{\sum_{i \in T} w_i}}}.

    \Avspace 
    \tcp{\textrm{$M$ here denotes the largest numerically stable number.}}
    \For{$i \in T$}{ 
      \lAssign{$\rho_i$}{\DS{
          \fOO
          {q_{ii} - w_i^{-1} \beta_\mu(T)}{w_i > 0}
          {M \sign\T{\beta_\mu(T)}}{w_i = 0}}}
    }
  }

  \Avspace

  \lIf{$\mrho_{[T]} = 0$} {
    \Return{$\emptyset$} \tcp{\textrm{These nodes are all at a common reduction level already.}}
  }
  \Avspace
  \Assign{$\mQ'$}{$\mQ_{[T]}$ with diagonal replaced by $\mrho_{[T]}$.}

  \Assign{$S^*_T$}{Minimum cut on $(T, \Ecal_{[T]})$, with capacities
    formed from $\mQ'$ by theorem \ref{th:equivalences}.
  }

  \Avspace 
  \Return{$S^*_T$}

  \label{algo:weighted-bisection}
\end{algorithm}

Note that what is important once we are done with the algorithm is not
the resulting values of $r_\mu(T)$, but that the edges spanning the
cut are saturated.  These are fixed per the same logic as in section
\ref{sec:algo}.

\newcommand\WBisectReductions{\textsc{WeightedBisectReductions}\xspace}
\newcommand\ApplyCut{\textsc{ApplyCut}\xspace}
\begin{algorithm}[thbp]
  \SetKw{Step}{step}
  \dontprintsemicolon
  \caption{\textsc{FindWeightedReductions}}
  \KwIn{Submodular $\mQ$, vector of non-negative weights $\mw$.}
  \KwOut{$\malpha^*$ satisfying condition \ref{eq:uawd}.}
  \Avspace
  
  \tcp{\textrm{\it Begin by calling \WBisectReductions below 
      on the full set $\Vcal$ to get $\malpha$.}}
  \Return{\WBisectReductions($T=\Vcal$, $\malpha = \m0$, $\mQ$)}
  \Avspace
  \Avspace
  \Avspace

  \tcp{\textrm{\it Note: $\malpha_{[T]}$ denotes $\malpha$ restricted to edges with both nodes in $T$.}}
  \Avspace 
  \WBisectReductions($T$, $\malpha$, $\mQ$)\;
  \Avspace 
  \Indp
  \Assign{$S^*_T$}{\textsc{WeightedBisectionCut}$(\mQ, \mw, T)$}
  \Avspace 

  \If{$S^*_T = \emptyset$ {\bf or } $S^*_T = T$}{
    \tcp{\textrm{We are done on this set.}}
    \Return{$\malpha_{[T]}$}
  }

  \Avspace 
  \tcp{\textrm{\it Fix the flow on edges in the cut by adjusting the
      source/sink capacities of each node, then removing those
      edges. }}

  \Avspace 
  \lFor{$i \in S_T^*,\, j \in T \backslash S_T^*,\, i<j$}{
    \lAssign{$\alpha_{ij}$}{$-q_{ij}$}
    \lAssign{$q_{ii}$}{$q_{ii} - q_{ij}$} \Assign{$q_{ij}$}{$0$}
  }
  \Avspace 
  \lFor{$i \in T \backslash S_T^*,\, j \in S_T^*,\,  i<j$}{
    \lAssign{$\alpha_{ij}$}{$q_{ij}$}
    \lAssign{$\;\;\,q_{jj}$}{$q_{jj} + q_{ij}$} \Assign{$q_{ij}$}{$0$}
  }

  \Avspace
  \tcp{\textrm{\it Recursively solve on the two partitions to fix the other $\malpha$'s. }}
  
  \Avspace
  \Assign{$\malpha_{[S_T^*]}$}{\WBisectReductions($S_T^*$, $\malpha$, $\mQ$)}
  \Assign{$\malpha_{[T \backslash S_T^*]}$}{\WBisectReductions(
    $T \backslash S_T^*$, $\malpha$, $\mQ$)}
  \Avspace

  \Return{$\malpha_{[T]}$}\;
  

  
  

  \label{algo:weighted-reduction}
\end{algorithm}
\begin{mathTheorem}[Correctness of Algorithm \ref{algo:weighted-reduction}]
  \label{th:algo-general-weights}
  Upon termination, algorithm \ref{algo:weighted-reduction} correctly
  finds $\malpha^*$.
\end{mathTheorem} %
\begin{proof}
  This proof is nearly identical to the proof of algorithm
  \ref{algo:reduction}; the difference is that we instead work with
  the condition of optimality given in corollary
  \ref{cor:opt-weighted-cut-solution:alt}.  

  Note first that for $w_i = 0$, the method used to set the values of
  $\rho_i$ on these nodes effectively sets them to their numerical
  extremes, ensuring that all operations seek to pull them as far as
  possible towards zero.  Thus we are consistent with the behavior of
  theorem \ref{th:zero-approximation}.

  Let $i,j$, $i < j$, be any pair of nodes such that $q_{ij} \neq 0$,
  and let $\malpha^\dagger$ be the solution returned by Algorithm
  \ref{algo:weighted-reduction}.
  
  First, suppose there existed a $\beta$ such that 
  \begin{align}
    r_j(\malpha^\dagger) - \beta w_j = r_i(\malpha^\dagger) - \beta w_i = 0
  \end{align}
  Then trivially, the optimality criterion of Lemma
  \ref{cor:opt-weighted-cut-solution:alt} is satisfied.

  Next, suppose that there exists a $\beta$ such that 
  \begin{align}
    r_i(\malpha^\dagger) - \beta w_i &\leq 0 < r_j(\malpha^\dagger) - \beta w_j \label{eq:e9934e} 
  \end{align}
  Then, according to the termination criterion of
  \textsc{WeightedBisectionCut} -- in which it returns $S^*_T = T$ or
  $S_T^* = \emptyset$ -- the edge between nodes $i$ and $j$ will be
  saturated in such a way that the reduction value of
  $r_i(\malpha^\dagger)$ is maximized and the reduction value of
  $r_j(\malpha^\dagger)$ is minimized, i.e. $\alpha_{ij} =
  -\absof{q_{ij}}$.  Thus condition \eqref{eq:joqoi:1} in corollary
  \ref{cor:opt-weighted-cut-solution:alt} is satisfied.  Similarly, if
  the order in \ref{eq:e9934e} is reversed, then condition
  \eqref{eq:joqoi:2} is satisfied.

  As the above holds for any pairs of nodes $i, j$, the optimality
  criteria of corollary \ref{cor:opt-weighted-cut-solution:alt} are
  satisfied globally, proving the correctness of the algorithm.
\end{proof}

\section{General Optimization with Unary Regularizers}
\label{sec:1181}

This section improves upon the result of theorem \ref{th:gen-opt-safe}
given at the end of section \ref{sec:ch:structure}: using the weighting
method above, it is possible to anchor weights at specific values in
the optimization.  This indicates that the reduction levels of these
nodes do not vary in the optimization even though they are treated the
same way as the other nodes.  This provides a simple way of
incorporating an $L_1$ fit penalty into our optimization.  Formally,
the following corollary to theorem \ref{th:gen-opt-safe} lays this
out:

\begin{mathTheorem}[]
  \label{th:gen-opt-safe-l1}
  Consider \eqref{eq:RB}: suppose $\gamma \ST \Reals^n \rightarrow
  \Reals^+$ can be expressed as
  \begin{equation}
    \gamma(\m u) = \const + \sqnorm{\m u - \m a}
    + \lambda\Tbr{\sum_i \xi_i(u_i)
      +  \sum_{i,j} w_{ij} \absof{u_i - u_j}}
    \label{eq:12jye}
  \end{equation}
  with $\ma, \mb \in \Reals^n$, $\lambda \geq 0$, and $w_{ij} \geq 0$,
  and where $\xi_i$ is a convex piecewise-linear function.  Then the
  minimizer
  \begin{equation}
    \m u^* = \argmin_{\m u \in \Reals^n} \gamma(\m u) 
    \label{eq:P0ynY}
  \end{equation}
  can be found exactly using Algorithm \ref{algo:weighted-reduction}.
\end{mathTheorem} %
\begin{proof}
  First, to accommodate the $\xi_i(u_i)$ functions, suppose, without
  loss of generality, that each function is defined by a sequence of
  $m_i$ inflection points
  \begin{equation}
    -\infty = b_{i0} < b_{i1} < b_{i2} < \cdots < b_{im_i} = \infty 
    \label{eq:sgSnP}
  \end{equation}
  and an associated line slope $\theta_{ij}$ between each, so 
  \begin{equation}
    \theta_{ij} = \frac{\xi_i(b_{ij}) - \xi_i(b_{i,j-1})}{b_{ij} - b_{i,j-1}}.
    \label{eq:ParkW}
  \end{equation}
  Then, since $\xi_i(x)$ is linear, it is easy to show that at any
  point $x$, $\xi_i(x)$ can be expressed as the integral over sums of
  step functions; if $c_i^-$ and $c_i^+$ are the values of these step
  functions,
  \begin{equation}
    \xi_i(u_i) = \text{const} + \int_{-M}^{u_i} 
    \Tcbr{\Tbr{\sum_{j : b_{ij} \leq x} c_{ij}^+} 
      + \Tbr{\sum_{j : b_{ij} > x'} c_{ij}^-}} \ud x.
    \label{eq:7hqfl}
  \end{equation}
  The exact values of $c_{ij}^+$ and $c_{ij}^-$ can be found easily by
  solving a simple linear system, and they can satisfy $c_{ij}^+ +
  c_{ij}^- = 0$ by adjusting the constant term in front of the
  integral; this ensures these pairwise terms satisfy the
  submodularity condition.  Thus $\xi_i(\cdot)$, can be encoded on the
  graph by creating $m_i$ auxiliary nodes $u'_{i,j}$ fixed at
  reduction values $b_{i1}, ..., b_{i,m_i}$ using the method described
  in section \ref{sec:915}.  Setting the overall weight of the edge
  using
  \begin{equation}
    E_2(x_i, x'_{ij}) = \Mccrr{0}{c_{ij}^+}{c_{ij}^-}{1}
    \label{eq:omgTj}
  \end{equation}
  ensures that the cost of a cut at any region between $b_{ij}$ and
  $b_{i,j+1}$ incurs cost 
  \begin{equation}
    \Tbr{\sum_{j' \leq j} c_{ij'}^+} + \Tbr{\sum_{j' > j} c_{ij'}^-}.
    \label{eq:u2QFR}
  \end{equation}
  Thus, in the final integral, the reduction level $u_i$ incurs the
  penalty $\text{const} + \xi(u_i)$.
  The proof then follows identically to theorem \ref{th:gen-opt-safe}
  with some minor modifications -- namely, for the auxiliary nodes, we
  are actually working with:
  \begin{equation}
    \indset{M\cdot u_{n+i} \leq M \cdot \beta}
    \label{eq:4n1QF}
  \end{equation}
  but this easily converts to 
  \begin{equation}
    \indset{u_{n+i} \leq \beta}. 
    \label{eq:9xyMi}
  \end{equation}
  As the $L_2$ fit terms on the auxiliary nodes become constant in the
  limit, the rest of the proof of theorem \ref{th:gen-opt-safe} goes
  through for this form as well, giving us the desired result.
\end{proof}

In this chapter,

\section{Conclusion}

In this work, we outlined the theoretical connections between network
flows, results in submodular optimization, and regularized regression
problems such as the graph-guided LASSO. We rigorously established the
submodular structure underlying the optimization of this problem.
This motivated several novel algorithms, and extended some of the
theory surrounding the minimum norm algorithm.  We extended the
existing theory of size-constrained submodular optimization first
proposed by \citep{nagano2011size} to the weighted case.  This
theoretical tool has several important consequences.  In the case of
network flows, it gives us the ability to make nodes more or less
affected by the optimization process.  This opens the door to the
general optimization problem given in theorem
\ref{th:gen-opt-safe-l1}.

In part 2, we extend these results to develop a full treatment of the
entire regularization path over all $\lambda$. In the second part of
this work, we will explore further implications of this theory, namely
a technique to algorithmically recycle solutions for network flows
through the use of the structure presented here.

\appendix

\section{Proofs}
\label{sec:str:proofs}

In this section, we give a number of the proofs needed for the
previous theorems; they are given here for readability.

\begin{proof}[Proof of Theorem \ref{th:equivalences}.]
\label{prf:equivalence}

To show the equivalence between $\Pcal_E(\beta)$ and $\Pcal_Q(\beta)$, it is sufficient to verify 
\begin{equation}
  \sum_{i \in \Vcal} (E_i(x_i) - \beta x_i) + \sum_{\substack{(i,j) \in \Ecal \\ i < j}} E_{ij}(x_i, x_j) 
  = \mx^T (\mQ - \diag(\beta)) \mx + \sum_i E_i(0) + \sum_{i<j\ST (i,j) \in \Ecal} E_{ij}(0,0)
  \label{eq:342p}
\end{equation}
for arbitrary $\mx \in \set{0,1}^n$. To simplify notation, assume that
$E_{ij}(x_i, x_j) = 0$ for all $(i, j) \notin \Ecal$. Then
  \begin{align}
    \sum_{i,j} &\mx^T (\mQ - \diag(\beta)) \mx + \sum_i E_i(0) + \sum_{i<j\ST (i,j) \in \Ecal} E_{ij}(0,0)\\
    &= \sum_i x_i (q_{ii} - \beta) + \sum_{i<j} q_{ij} x_i x_j + \sum_i E_i(0) + \sum_{i<j\ST (i,j) \in \Ecal} E_{ij}(0,0)\\ 
    &= \sum_i \Tbrss{x_i (\E_i(1) - \beta - E_i(0)) + E_i(0)} 
    \EAsplithere{2em}
    + \sum_{i < j} \Tbrss{ x_ix_j\T{E_{ij}(1,1) + E_{ij}(0,0) - E_{ij}(0,1) - E_{ij}(1,0)} \EAsplithere{5em}
      + x_i\T{E_{ij}(1,0) - E_{ij}(0,0)} + x_j \T{E_{ij}(0,1) - E_{ij}(0,0)}} 
    \EAsplithere{2em}
    + \sum_{i < j} E_{ij}(0,0) \\
    &=
    \sum_i \fOOdbl{E_i(1) - \beta}{x_i = 1} {E_i(0)}{x_i = 0}
    + \sum_{i < j} 
    \fOOOOdbl
    {E_{ij}(1,1)}{x_i = x_j = 1}
    {E_{ij}(1,0)}{x_i = 1, x_j = 0}
    {E_{ij}(0,1)}{x_i = 0, x_j = 1}
    {E_{ij}(0,0)}{x_i = 0, x_j = 0} \\ 
    &= \sum_{i \in \Vcal} (E_i(x_i) - \beta x_i) + \sum_{\substack{(i,j) \in \Ecal \\ i < j}} E_{ij}(x_i, x_j).
  \end{align}
  Thus the values of the objective function differ by a constant amount
  for all values $\mx$, so the set of minimizers is identical.  

  Now, to verify the equivalence of $\Pcal_Q(\beta)$ and
  $\Pcal_N(\beta)$, let $x_i$ be the associated indicator vector,
  given by:
  \begin{equation}
    x_i = \Ind{i \in \T{S \backslash \set{s} }} \quad \Leftrightarrow \quad S = \Set{i}{x_i = 1}
    \label{eq:I0Ww}
  \end{equation}
  It then suffices to verify that 
  \begin{equation}
    \sum_{(i,j) \in \delta(S(\beta), \Vcal \backslash S(\beta)) } c_{ij} = \mx(\beta)^T (\mQ - \diag(\beta)) \mx(\beta) + \sum_i c_{i,t}
  \label{eq:G9w9}
\end{equation}

for all $\mx \in \set{0,1}^n$ and corresponding cuts $\set{s} \subseteq S \subseteq \Vcal'$:
\begin{align}
  \operatorname{Cost}(S, \Vcal \backslash S) &=
  \sum_{i \in S, j \in T} c_{ij} + \sum_{i \in T} c_{si} + \sum_{i \in S} c_{it} \\
  &=
  \Tbr{\sum_{i,j \ST j \in T} c_{ij} - \sum_{i,j \in T} c_{ij}} + \sum_{i \in T} (c_{si} - c_{it}) + \sum_i c_{it} \\
  &= 
  \Tbr{\sum_{i' < i \ST i \in T} c_{i',i}+ \sum_{i < j \ST i \in T} c_{ij}}
  + \sum_{i \in T} (c_{si} - c_{ti}) \EAsplithere{3em}
  + 2 \sum_{i < j\ST i,j \in T} \T{- c_{ij}}
  + \sum_i c_{it} \\
  &= 
  \sum_{i' < i \ST i \in T} \T{- \inv{2} q_{i',i} } + \sum_{i < j \ST i \in T} \T{- \inv{2} q_{ij}} \EAsplithere{1em}
  + \sum_{i \in T} \Tbr{q_{ii} + \inv{2}\sum_{i'\ST i' < i} q_{i',i} + \inv{2}\sum_{j \ST i < j} q_{ij}} 
  + 2 \sum_{i < j\ST i,j \in T} \inv{2} q_{ij}
  + \sum_i c_{it} \\
  &= 
  \mx^T (\mQ - \diag(\beta) ) \mx + \sum_i c_{it}.
\end{align}
Again, the values of the objective functions differ by a constant
amount for all $\mx$, proving the theorem.
\end{proof}

\begin{proof}[Proof of Theorem \ref{th:is-submodular}.]

  From \citep{fujishige2005submodular}, we know that $f_\beta$ is
  submodular if and only if, for all $T \subset U$ and $v \in S
  \backslash U$,
  \begin{equation}
    f_\beta(T \cup \set{v}) - f_\beta(T) \geq f_\beta(U \cup \set{v}) - f_\beta(U).
    \label{eq:npTg}
  \end{equation}
  For convenience, assume that the $\beta$ is absorbed into the diagonal elements of $q_{ii}$.
  Now 
  \begin{align}
    \Tbr{f_\beta(T \cup \set{v}) - f_\beta(T)} - &\Tbr{f_\beta(U \cup \set{v}) - f_\beta(U)} \\
    &= \Tbr{\sum_{i,j \in T} q_{ij} + \sum_{i \in T} (q_{iv} + q_{vi}) + q_{vv} - \sum_{i,j \in T} q_{ij}}
    \EAsplithere{5em}
    - \Tbr{\sum_{i,j \in U} q_{ij} + \sum_{i \in U} (q_{iv} + q_{vi}) + q_{vv} - \sum_{i,j \in U} q_{ij}} \\
    &= \Tbr{\sum_{i \in T} (q_{iv} + q_{vi})} - \Tbr{\sum_{i \in U} (q_{iv} + q_{vi})} \\
    &= - \sum_{i \in U\backslash T} (q_{iv} + q_{vi}). \label{eq:1ed3}
  \end{align}
  The theorem immediately follows; this is non-negative if $q_{ij}
  \leq 0\Forall i < j$, and $T, U, v$ can be easily chosen to make
  it negative if $\Exists i,j$ such that $q_{ij} > 0$.
\end{proof}

\begin{proof}[Proof of theorem \ref{th:polymatroid}.]

  For ease of notation, assume that $\beta w_i$ is absorbed into
  $q_{ii}$. For ease of algebra, we first prove a simpler result, then
  show it immediately implies the desired one. 

  Define
  \begin{equation}
    \Wcal = \Set{\mw \in \Mcal_{n \times n} }{
      \fOO
      {q_{ij} \leq w_{ij} \leq 0}{\text{if } i < j} 
      {w_{ij} = 0}{\text{otherwise}}}.
    \label{eq:353}
  \end{equation}
  (Recall that $q_{ij} \leq 0$.) Then, we show that the associated
  polymatroid $P(f_\beta)$ can be represented by
  \begin{equation}
    P(f_\beta) = \Set{\my}{ \Exists \mw \in \Wcal 
      \st \Forall i, \; y_i  \leq q_i 
      + \sum_{i'\ST i' < i} w_{i',i} 
      + \sum_{j\ST i < j} (q_{ij} - w_{ij})}.
    \label{eq:342}
  \end{equation}
  To show that $P(f_\beta)$ is the base of $f$, we must show that
  $\forall\;\my \in P(f_\beta)$ and $\forall\;U \subseteq \Vcal$, $\sum_{i
    \in U} y_i \leq f(U)$ and that $\forall\;U\subseteq\Vcal$,
  there exists a point $\my \in P(f_\beta)$ such that $\sum_{i\in U} y_i =
  f(U)$ \citep{schrijver2003combinatorial}.

  To show equation (\ref{eq:342}), we first prove that $\Forall \my
  \in P(f_\beta)$, $\sum_{i\in U} y_i \leq f(U) \Forall U \subseteq S$.  Now
  \begin{align}
    \sum_{i \in U} y_i 
    &\leq \sum_{i \in U} \Tbr{q_{ii} 
      + \sum_{i'\ST i' < i} w_{i',i} 
      + \sum_{j\ST i < j} (q_{ij} - w_{ij})} \label{eq:2eie:2} \\
    &\leq \sum_{i \in U} \Tbr{q_{ii}
      + \sum_{i' \in U \ST  i' < i} w_{i',i} 
      + \sum_{j \in U \ST i < j} (q_{ij} - w_{ij})} \label{eq:2eie:3} \\
    &= \sum_{i \in U} q_{ii}  + \sum_{i,j \in U} q_{ij} 
    = f(U)
  \end{align}
  It remains to show that $\Forall U \subseteq \Vcal$, $\Exists \my \in
  P(f_\beta)$ such that $\sum_{i \in U} y_i = f(U)$.  For this, set 
  \begin{equation}
    w^*_{ij} = \fOO{q_{ij}}{i \in U,\, i < j}{0}{\text{otherwise}}
    \label{eq:379}
  \end{equation}
  Clearly, $\mw^* \in \Wcal$. Setting $\my^*$ to be the extreme point
  of $P(f_\beta)$ using $\mw = \mw^*$, we have, for all $U \subseteq \Vcal$,
  \begin{align}
    \sum_{i \in U} y^*_i 
    = \sum_{i \in U} \Tbr{q_{ii} 
      + \sum_{i' \ST i' < i} w^*_{i',i} 
      + \sum_{j \ST i < j} (q_{ij} - w_{ij})}
    = \sum_{i \in U} q_{ii} 
      + \sum_{i', i \in U \ST i' < i} q_{i',i}  = f(U)
  \end{align}
  Thus \eqref{eq:342} defines the polymatroid polytope of $f_\beta$.
  Similarly, we show that the base of $P(f_\beta)$ can be given by
  \begin{equation}
    B(f_\beta) = \Set{\my}{ \Exists \mw \in \Wcal 
      \st \Forall i,\;y_i = q_{ii} + \sum_{i'\ST i' < i} w_{i',i} 
      + \sum_{j\ST i < j} (q_{ij} - w_{ij})}
    \label{eq:366}
  \end{equation}
  To show this, note that, trivially, $B(f_\beta) \subset P(f_\beta)$.
  It remains to show that for all $\my \in B(f_\beta)$, $\sum_i y_i =
  f(\Vcal)$ for all $\mw \in \Wcal$:
  \begin{align}
    \sum_i y_i = \sum_i q_{ii} 
    + \sum_{i',i \ST i' < i} w_{ij} 
    + \sum_{i, j \ST i < j} (q_{ij} - w_{ij}) 
    = \sum_i q_{ii} + \sum_{i,j \ST i<j} q_{ij} 
    = f(S)
  \end{align}
  Finally, the theorem follows by replacing $w_{ij}$ with $\onehalf
  (\alpha_{ij} - q_{ij})$ in the above results and bringing the
  $\beta w_i$ outside of $q_{ii}$ and $\mr(\malpha)$.
\end{proof}

\begin{proof}[Proof of theorem \ref{th:gen-opt-safe}]
\label{th:gen-opt-safe:proof}
  First, we can write $\gamma(\m u)$ as 
  \begin{align}
    \gamma(\m u) 
    &= \inv\lambda \m u^T \m u - \frac{2}{\lambda} \m a^T \m u + \sum_{i,j} w_{ij} \absof{u_i - u_j} + \const \\
    &\propto \inv{2} \m u^T \m u - \m a^T \m u + \sum_{i,j} \frac{\lambda w_{ij} }{2} \absof{u_i - u_j} + \const 
    \label{eq:rSHHb}
  \end{align}
  Now, define $\indset{\m u \leq \beta}$ as the cut vector of $\m u$
  at $\beta$, given by
  \begin{equation}
    \indset{\m u \leq \beta} = \T{\indset{u_i \leq \beta} \ST i = 1,2,...,n} \subseteq \set{0,1}^n,
    \label{eq:gAZcD}
  \end{equation}
  Now, assume that $u_i \in \Icc{-M, M}$ for $i = 1,2,...,n$, where $M$ is
  sufficiently large.  This is reasonable, as $\lambda^{-1} \m u^T \m u$
  dominates the optimization asymptotically.  This enables us to write
  the terms in the above expression as
  \begin{align}
    \inv{2} \m u^T \m u = \inv{2}\sum_{i=1}^n u_i^2 &= \const + \sum_{i = 1}^n \int_{-M}^M \beta\indset{\beta < u_i} \ud \beta \label{eq:qkmwj:1} \\
    &= \const + \sum_{i = 1}^n \int_{-M}^M (-\beta) \indset{u_i \leq \beta} \ud \beta \label{eq:qkmwj:1b} \\
    -\ma^T \m u 
    &= \text{const} - \sum_{i = 1}^ n \int_{-M}^M \ma^T\indset{\beta < u_i} \ud \beta \label{eq:qkmwj:2} \\
    &= \text{const} + \sum_{i = 1}^ n \int_{-M}^M \ma^T\indset{u_i \leq \beta} \ud \beta \label{eq:qkmwj:2b} \\
    \sum_{i,j} \frac{\lambda w_{ij}}{2} \absof{u_i - u_j} 
    &= \int_{-M}^M \sum_{i,j} \frac{\lambda w_{ij}}{2}
    \Indset{ \indset{u_i \leq \beta} \neq \indset{u_j \leq \beta}} \ud \beta \label{eq:qkmwj:3}.
  \end{align}
  Putting this together gives us
  \begin{equation}
    \gamma(\m u) \propto \text{const} + 
    \sum_i \int_{-M}^M \Tcbr{ \Tbr{(\ma - \beta)^T\T{\indset{\m u \leq \beta} }
      + \Tbr{\sum_{i,j} \frac{\lambda w_{ij}}{2}
        \Indset{\indset{u_i \leq \beta} \neq \indset{u_j \leq \beta}}}}} \ud \beta
    \label{eq:MkNTv}
  \end{equation}
  However, the first term in the given approach maps directly to the
  unary $q_{ii} - \beta$ terms in the graph problem earlier, with the
  $\indset{\m u \leq \beta}$ term indexing the cut from source to node
  or from node to sink.  Thus, for a given $\beta$, $\indset{u_i \leq
    \beta}$ is a binary variable with $\indset{u_i \leq \beta}$
  indicating that the edge from source to node $i$ is cut, incurring
  the appropriate cost in the overall function.

  Similarly, $\Indset{ \indset{u_i \leq \beta} \neq \indset{u_j \leq
      \beta}}$ is 1 for the range of $\beta$ in which the edge
  connecting nodes $i$ and $j$ is cut; i.e. where the indicator
  variables of set membership differ.  The cost of cutting this edge
  is given by $\onehalf\lambda w_{ij}$, and it is incurred by all
  points in the range of $\beta$ between $\min(u_i, u_j)$ and
  $\max(u_i, u_j)$.  Integrating this gives us $\onehalf\lambda w_{ij}
  \absof{u_i - u_j}$.  This establishes \eqref{eq:MkNTv}.
  
  Now, it remains to show that finding the minimum energy cut solution
  of the above problem at all $\beta$ is equivalent to finding $\m
  u^*$.  For this step, define a set of functions mapping $\Reals$ to
  $\set{0,1}$ as:
  \begin{align}
    \Sscr = \Set{\Indset{\,\cdot\, \leq t}}{t \in \Reals}.
    \label{eq:6dNpL}
  \end{align}
  Now, we can rewrite the minimization of $\gamma(\m u)$ as
  \begin{align}
    &\min_{u \in \Icc{-M, M}^n} \gamma(\m u)  \\
    &\quad\propto \const + 
    \min_{u \in \Icc{-M, M}^n}
    \int_{-M}^M \Tcbr{ \Tbr{(\ma - \beta)^T\T{\indset{\m u \leq \beta} }
      + \Tbr{\sum_{i,j} \frac{\lambda w_{ij}}{2}
        \Indset{\indset{u_i \leq \beta} \neq \indset{u_j \leq \beta}}}}} \ud \beta \label{eq:1joyie:1} \\
    &\quad= 
    \const + 
    \min_{\ms \in \Sscr^n} 
    \int_{-M}^M \Tcbr{ \Tbr{(\m a - \beta)^T \ms(\beta)}
      + \Tbr{\sum_{i,j} \frac{\lambda w_{ij}}{2} \Indset{ s_i(\beta) \neq s_j(\beta)}}} \ud \beta \label{eq:1joyie:2}
  \end{align}
  where $\ms = s_1, s_2, ..., s_n$, $s_i \in \Sscr$ replaces the
  optimization over the changepoints $u_i$.  For convenience, denote
  the term in the integral by
  \begin{equation}
    h(\beta, \mx) = \Tbr{(\m a - \beta)^T \mx}
      + \Tbr{\sum_{i,j} \frac{\lambda w_{ij}}{2} \Indset{x_i \neq x_j}}
    \label{eq:kKN9G}
  \end{equation}
  where $\mx \in \set{0,1}^n$.  Now it is easy to see that
  \begin{align}
    \min_{\ms \in \Sscr^n} 
    \int_{-M}^M h(\beta, \ms(\beta)) \ud \beta  
    \geq
    \int_{-M}^M \Tcbr{\min_{\mx \in \set{0,1}^n} h(\beta, \mx) } \ud \beta.
  \end{align}
  However, by corollary \ref{cor:nestedness}, we know that for any
  $\beta_1 < \beta_2$, with
  \begin{align}
    \mx^*(\beta_1) &\in \Argmin_{\mx \in \set{0,1}^n} h(\beta_1, \mx) \\
    \mx^*(\beta_2) &\in \Argmin_{\mx \in \set{0,1}^n} h(\beta_2, \mx),
    \label{eq:4mWWu}
  \end{align}
  we have the following implications for all $i$:
  \begin{align}
    \text{if } x^*_i(\beta_1) = 0, \quad &\text{ then } \quad x^*_i(\beta_2) = 0 \\
    \text{if } x^*_i(\beta_2) = 1, \quad &\text{ then } \quad x^*_i(\beta_1) = 1.
  \end{align}
  This, however, is exactly the constraint implied in the set $\Sscr$.
  Thus \eqref{eq:4mWWu} holds with equality:
  \begin{equation}
    \min_{\ms \in \Sscr^n} 
    \int_{-M}^M h(\beta, \ms(\beta)) \ud \beta  
    =
    \int_{-M}^M \Tcbr{\min_{\mx \in \set{0,1}^n} h(\beta, \mx) } \ud \beta.
    \label{eq:zNBxN}
  \end{equation}
  However, this minimization is exactly the problem solved earlier;
  namely, by theorem \ref{th:opt-cut-solution}, we find the optimal
  set for each $\beta$.  Thus, the net result of this operation is
  that $\m u^* = \mr(\malpha^*)$.  Letting $M \goesto \infty$
  completes the theorem.
\end{proof}

\begin{proof}[Proof of lemma \ref{lem:integer-weight-addition}]
\label{lem:integer-weight-addition:proof}  %
  \nid ({\bf \ref{lem:augment:1}})
  First, we establish that $f_\mw$ is indeed a
  submodular function defined on a distributed lattice.  As given,
  $f_\mw$ is defined on
  \begin{align}
    \Dcal' 
    &= \Set{T \in 2^{\Vcal_\mw}}{T \cap \Vcal \in \Dcal}. \label{eq:u94E} \\ 
    &= \Set{S \cup U}{S \in \Dcal,\; U \in 2^{\Vcal_\mw \backslash \Vcal}}. 
  \end{align}
  This is a distributed lattice, as it is the product of the
  distributed lattices $\Dcal$ and $2^{\Vcal_\mw \backslash \Vcal}$.
  Now, let 
  \begin{equation}
    g_{ij}(U) = M\cdot\T{\indset{i \in T} + \indset{j \in T} - 2 \ind{ \set{i, j} \subseteq T}}
    \label{eq:MRwe}
  \end{equation}
  for $i \in \Vcal$ and $j \in \Vcal_\mw \backslash \Vcal$. It is easy
  to see that $g_{ij}$ is a submodular function, as 
  \begin{equation}
    2M = g(\set{i}) + g(\set{j}) \geq g(\set{i,j}) + g(\emptyset) = 0
    \label{eq:t58q}
  \end{equation}
  for any $i,j$.  Thus $f_\mw(T)$ is a submodular function on
  $\Dcal'$, as it is the sum of $W + 1$ submodular functions.
  
  Now, we establish \eqref{eq:owVE} inductively. Define a sequence of
  $W$ integers $k_1,k_2,..., k_W$ satisfying
  \begin{align}
    w_i = \#\Set{j}{k_j = w_i} + 1 \Forall i \in \Vcal. 
  \end{align}
  Such a set may be chosen by reversing $K_i$ in \eqref{eq:nXK6} above. 

  Let $\Vcal_m = \Vcal \cup \set{n+1, ..., n+m}$.  Define $f_{0, \beta}(S) =
  f'_{0,\beta}(S) = f(S) - \beta\sizeof{S}$, and let
  \begin{align}
    f_{m,\beta}(S) &= f_{m-1,\beta} (S) - \beta \indset{k_j \in S}.
    & S &\subseteq \Vcal 
    \label{eq:BP5w1} \\
    f'_{m,\beta}(T) &= f'_{m-1,\beta}\T{T \cap \Vcal_{m-1}} + g_{k_m,n+m}(T) - \beta\indset{n + m \in T}, & T &\subseteq \Vcal_m 
    \label{eq:BP5w2}
  \end{align}
  We intend to show that for $m = 1,..., W$,
  \begin{equation}
    \Argmin_{S \subseteq \Dcal} f_{m,\beta}(S) + h(S) 
    = \Set{\Vcal \cap T^*}
    {T^* \in \Tbr{\Argmin_{T \subseteq \Vcal_m \ST T\cap \Vcal \in \Dcal} f'_{m,\beta}(T) + h(T \cap \Vcal)}}
    \label{eq:1Pmu}
  \end{equation}
  for all modular functions $h$ on $\Vcal$. 

  Fix $\beta$ and $h$. Trivially, \eqref{eq:1Pmu} is true for the
  ground case $m = 0$.  Now suppose \eqref{eq:1Pmu} is true for $m-1$,
  and let
  \begin{align}
    T^* &\in \Argmin_{T \subseteq \Vcal_m \ST T\cap \Vcal \in \Dcal} f'_{m,\beta}(T) + h(T)
  \end{align}
  be any minimizer of $f'_{m,\beta}$.  

  To show that the elements $k_m$ and $n+m$ are tied in $T^*$, assume
  the opposite -- suppose that either $\Tbr{k_m \in T^* \text{ and }
    n+m \notin T^*}$ or $\Tbr{k \notin T^* \text{ and } n+m \in T^*}$.
  This, however, contradicts the optimality of $T^*$ for sufficiently
  large $M$, as the value of the minimum can always be improved by $M$
  if element $n+m$ is included or excluded so that its membership in
  $T^*$ matches that of $k_m$.  Thus, for optimal $T^*$, $k_m \in T^*$
  if and only if ${n+m \in T^*}$. 

  We then have that for any minimizer $T^*$ of $f'_{m,\beta} + h$,
  \begin{equation}
    f'_{m,\beta}(T^*) = f'_{m-1,\beta}\T{T^* \cap \Vcal_{m-1}} - \beta\indset{k_m \in T^*}, 
    \label{eq:lIdo}
  \end{equation}
  as $g_{k_m,n+m}(\set{k_m, n+m}) = g_{k_m,n+m}(\emptyset) = 0$.

  Now let $h'(S) = h(S) - \beta\indset{k_m \in T^*}$.  Then
  \begin{align}
    \Argmin_{S \subseteq \Dcal} &f_{m,\beta}(S) + h(S) \label{eq:euom0} \\ 
    &= \Argmin_{S \subseteq \Dcal} f_{m-1,\beta}(S) + h'(S)  \label{eq:euom1} \\
    &\quad= \Set{\Vcal \cap T^*}
    {T^* \in \Tbr{\Argmin_{T \subseteq \Vcal_{m-1} \ST T \cap \Vcal \in \Dcal} f'_{m-1,\beta}(T) + h'(T\cap \Vcal)}}
    \label{eq:euom2} \\
    &\quad= \Set{\Vcal \cap T^*}
    {T^* \in \Tbr{\Argmin_{T \subseteq \Vcal_m \ST T \cap \Vcal \in \Dcal} f'_{m,\beta}(T) + h(T \cap \Vcal)}}
    \label{eq:euom3}
  \end{align}
  where \eqref{eq:euom1} - \eqref{eq:euom2} follows by the inductive
  assumption.  As \eqref{eq:euom1} - \eqref{eq:euom2} holds for any
  modular function $h'(S)$, we have that \eqref{eq:euom0} -
  \eqref{eq:euom3} is true as well for any $h(S)$, proving
  \eqref{eq:1Pmu} for $m = 1, 2, ..., W$.

  Now, it is easy to show that 
  \begin{align}
    f'_{W,\beta}(T) 
    &= f(T \cap \Vcal) + \sum_{i = 1}^{W} g_{k_i,n+i}(T) - \beta \sizeof{T}  \\
    &= f_\mw(T) - \beta\sizeof{T} 
  \end{align}
  For all $T \in \Dcal'$, and
  \begin{align}
    f_{W,\beta}(S) 
    &= f(S) - \beta \sum_{i \in \Vcal} w_i \indset{i \in S} \\
    &= f(S) - \beta \mw(S)
  \end{align}
  for all $S \in \Dcal$, proving the first part of the theorem.
  
  \nid ({\bf \ref{lem:augment:2}})
  To show \ref{lem:augment:2}, note that the
  membership of each item in $\Vcal_\mw \backslash \Vcal$ exactly
  matches the membership of an item in $\Vcal$; as $\Dcal$ is a
  distributed lattice, $\Dcal_\mw$ is also a distributed lattice that
  is closed under intersection and union.

  It remains to show that \eqref{eq:owVE2} is equivalent to
  \eqref{eq:owVE}.  This follows directly from noting that for $M$
  sufficiently large, $\Dcal_\mw$ can be written as
  \begin{equation}
    \Dcal_\mw = \Set{T \subseteq \Vcal_\mw}{T \cap \Vcal \in \Dcal,\; f_\mw(T) \leq M / 2}.
    \label{eq:luBC}
  \end{equation}
  As we have already argued that in any optimal solution $T^*$ of
  $f_\mw(T)$, the $M$ terms cancel out, so any minimizer of $f_\mw(T)$
  must be in $\Dcal_\mw$, proving \ref{lem:augment:2}.
  
  \nid ({\bf \ref{lem:augment:3}}) Finally, in this case, for all
  $\set{j,k}$ such that $\Exists S \in \Dcal_\mw$ with $\set{j,k}
  \subseteq S$, $g_{jk}(S) = 0$, proving \eqref{eq:asw2} and
  completing the proof.
\end{proof}

\begin{proof}[Proof of theorem \ref{th:min-norm-stuff}]
  \label{th:min-norm-stuff:proof}

  ({\bf \ref{th:min-norm-stuff:1}})
  Recall that the definition of the polymatroid associated with the
  submodular function $f_\mw$ is defined as
  \begin{equation}
    P(f_\mw) = \Set{\mx \in \Reals^{\sizeof{\Vcal_\mw}}}{\Forall T \in \Dcal_\mw, \mx(T) \leq f_\mw(T)}
    \label{eq:XY4P}
  \end{equation}
  Now, by equation \eqref{eq:asw2} in lemma \ref{lem:integer-weight-addition}, 
  \begin{equation}
    f_\mw(T) = f(T \cap \Vcal) 
    \label{eq:KN72}
  \end{equation}
  for all $T \in \Dcal_\mw$. Furthermore, for all $j \in K_i$, $i \in
  T$ if and only if $j \in T$.  Thus the condition $\mx(T) \leq
  f_\mw(T)$ is equivalent to 
  \begin{align}
    \sum_{i \in T \cap \Vcal} \mx\T{\kappa_i} &\leq f(T \cap \Vcal)
    \label{eq:CNZ0}
  \end{align}
  and the condition $\mx(\Vcal_\mw) = f_\mw(\Vcal_\mw)$ is equivalent
  to
  \begin{equation}
    \mx(\Vcal_\mw) = \sum_{i \in \Vcal} \mx\T{\kappa_i} = f(\Vcal).
    \label{eq:B2mf}
  \end{equation}
  This proves part \ref{th:min-norm-stuff:1}

  \nid ({\bf \ref{th:min-norm-stuff:2}-\ref{th:min-norm-stuff:3}})
  Now, to prove part \ref{th:min-norm-stuff:2}, consider minimizing
  $\lnorm{\my}{2}^2$ over $B(f_\mw)$.  Under the constraints, we can
  express this problem as 
  \begin{align}
    (\star) \quad \minimize&\qquad \sum_{i \in \Vcal} \Tbr{\sum_{j \in \kappa_i} y_j^2}  \nonumber \\
    \text{such that}&\qquad \sum_{i \in S} \my\T{\kappa_i} \leq f(S) \Forall S \in \Dcal \label{eq:jpa2}\\
    &\qquad \sum_{i \in \Vcal} \my\T{\kappa_i} = f(\Vcal) \nonumber
  \end{align}
  Now, let $z_i = \my\T{\kappa_i}$, and define 
  \begin{equation}
    h_i(\my, t) = \min\Set{\sum_{j \in \kappa_i} y_j^2}{ \sum_{j \in \kappa_i} y_j = t }
    \label{eq:EjSN}
  \end{equation}
  The objective $(\star)$ in \eqref{eq:jpa2} above can then be expressed as 
  \begin{equation}
    (\star) = \minimize \sum_{i \in \Vcal} h_i\T{\my, \my\T{\kappa_i}}.
    \label{eq:dgtr}
  \end{equation}
  This, however, has an analytic solution, as the $L_2$-norm above is
  minimized when $\my$ is equal on the set $\kappa_i$, i.e.
  \begin{equation}
    y_{j} = \frac{t}{\sizeof{\kappa_i}} \Forall j \in \kappa_i.
    \label{eq:d0gv}
  \end{equation}
  Thus
  \begin{equation}
    h_i(\my, t) 
    = h_i(t) 
    = \sizeof{\kappa_i} \T{\frac{t}{\sizeof{\kappa_i}}}^2 
    = \frac{t^2}{\sizeof{\kappa_i}} = \frac{t^2}{w_i}.
    \label{eq:j3Ke}
  \end{equation}
  This proves equation \eqref{eq:8mWg}, and allows us to rewrite
  \eqref{eq:jpa2} as
  \begin{align}
    \minimize&\qquad \sum_{i \in \Vcal} z_i^2 / w_i \nonumber \\
    \text{such that}& \qquad \mz(S) \leq f(S) \Forall S \in \Dcal \label{eq:jpa3}\\
    &\qquad \mz(\Vcal) = f(\Vcal) \nonumber
  \end{align}
  which is exactly identical to the optimization in equation
  \eqref{eq:ekbU}.  Given that $\my$ is constant on each $\kappa_i$,
  we have already proved equation \eqref{eq:CDfN} by setting $t = z_i$
  in equation \eqref{eq:d0gv}.
\end{proof}

\subsection{Proof of Correctness for the Weighted Minimum Norm Problem}
\label{sec:proof-wmn}

Before proving theorem \ref{th:general-weights}, we must establish a
lemma that shows convergence of the mapping in \eqref{eq:SIyy} under
various perturbations.  The reason that this lemma is needed is that
the crux of the proof of theorem \ref{th:general-weights} involves
showing that all positive $\mw \in \Reals^n$ can be seen as the limit
of a sequence of problems with integer $\mw \in \Zbb^n$.  The crux of
this lemma is found in \cite{wets2003lipschitz}, where a convenient
theorem shows that mappings of the type \eqref{eq:SIyy} are locally
Lipshitz-continuous with respect to perturbations of the objective
function.  This continuity is sufficient to ensure that the limiting
argument employed in the proof of theorem \ref{th:general-weights} is
valid.

\begin{mathLemma}[]
  \label{lem:lipshitz}
  Let $D$ be a convex set, and let 
  \begin{equation}
    g^*(\mdelta, \mw) = \argmin_{\mz \in D} \sum_{i = 1}^n \frac{(z_i + \delta_i)^2}{w_i}
    \label{eq:ke5O2}
  \end{equation}
  for $\mdelta, \mw \in \Reals^n$, with $\mw \geq \eta > 0$ bounded
  away from $0$.  Then $g^*(\mdelta, \mw)$ is a locally
  Lipshitz-continuous function of $\mdelta$ and $\mw$.  In other
  words, for every point $(\mdelta, \mw) \in \Reals^n \times
  \Ico{\eta, \infty}^n$, there exists a neighborhood $U \subset
  (\mdelta, \mw)$, $(\mdelta, \mw) \in U$, such that $g^*$ is
  Lipshitz-continuous on $U$.
\end{mathLemma} %
\begin{proof}
  Define $h \ST \Reals^n \times \Reals^n \times \Reals^n \mapsto
  \Reals \cap \set{\infty}$ as an extended version of the function $g$:
  \begin{equation}
    h(\mdelta, \mw, \mz) = \fOO
    {\sum_{i = 1}^n \frac{(z_i + \delta_1)^2}{w_i}}{\mz \in B(f)}
    {\infty}{\text{otherwise}}.
    \label{eq:epvRU}
  \end{equation}
  The lemma follows from showing that $h$ satisfies the properties of
  Theorem 3.4 of \citet{wets2003lipschitz}, which states that the
  $inf$-mapping of \eqref{eq:ke5O2} is locally Lipshitz continuous if
  mild conditions on $h$ are satisfied.  These conditions follow
  immediately as $h$ is Lipshitz continuous in $\mw$, $\mdelta$, and
  $\mz$, and the domain of $\mz$ is not affected by the $\mw$ and
  $\mdelta$.  This gives us the desired result.
\end{proof}

\begin{proof}[Proof of theorem \ref{th:general-weights}.]

\nid With this lemma in place, we are now ready to provide the proof
of theorem \ref{th:general-weights}.  The proof proceeds in two
stages.  First, we show that it is valid for positive rational $\mw
\in \Qbb^n_{++}$, where $\Qbb_{++} = \Set{x \in \Qbb}{x > 0}$.  Then,
we use a carefully constructed limiting argument extends this to all
real weights.

$ $ 

  \nid {\bf Step 1:} First, assume that the weights are
  positive rationals. Thus we may write
  \begin{equation}
    w_i = \frac{N_i}{M_i},
    \label{eq:lZ5o4}
  \end{equation}
  where $N_i \in \Zbb_{++}$ and $M_i \in \Zbb_{++}$. Then, let
  \begin{equation}
    M = \prod_{i \in \Vcal} M_i,
    \label{eq:dqAIL}
  \end{equation}
  so $M w_i$ is an integer for all $i \in \Vcal$.  Now note that we
  can immediately map the original problem to this form by simply
  setting 
  \begin{align}
    w'_i &= M w_i \\
    \beta' &= \beta / M
  \end{align}
  Then $\beta w_i = \beta' w_i'$, with $w'_i$ being an integer. 

  Let $\Vcal_{\mw'}$, $\Dcal_{\mw'}$, and $f_{\mw'}$ be as defined in theorem
  \ref{th:min-norm-stuff}, and let
  \begin{equation}
    \mz^* = \argmin_{\mz \in B(f)} \sum_{i \in \Vcal} \frac{z_i^2}{w_i} 
    = \argmin_{\mz \in B(f)} \inv{M} \sum_{i \in \Vcal} \frac{z_i^2}{w_i} 
    = \argmin_{\mz \in B(f)} \sum_{i \in \Vcal} \frac{z_i^2}{w'_i}
    \label{eq:FBxr2}
  \end{equation}
  and let
  \begin{equation}
    \my^* = \frac{\mz^*}{\mw'}.
    \label{eq:oYYnx}
  \end{equation}
  By theorem \ref{th:min-norm-stuff}, we know that $\mz^*$ is the optimal solution to
  \eqref{eq:FBxr2} if and only if $\my^*$ is the optimal solution to
  \begin{equation}
    \my^* = \argmin_{\my \in B(f_{\mw'})} \sqnorm{\my}.
    \label{eq:GNNt}
  \end{equation}
  Now, by theorem \ref{th:min-norm-stuff}-\ref{th:min-norm-stuff:3}
  and theorem \ref{th:min-norm-sfm}-\ref{th:min-norm-sfm:1}, we have
  that $\my^*$ is optimal for \eqref{eq:GNNt} if and only if
  $\forall\, \beta' \in \Reals$, all optimal solutions
  \begin{equation}
    T^*(\beta') = \argmin_{T \in \Dcal_{\mw'}} (f_{\mw'}(T) - \beta')
    \label{eq:UGpy}
  \end{equation}
  satisfy
  \begin{equation}
    U'_1(\beta') = \Set{i \in \Vcal}{y^*_i - \beta' < 0} 
    \subseteq T^*(\beta')
    \subseteq \Set{i \in \Vcal}{y^*_i - \beta' \leq 0} = U'_2(\beta').
    \label{eq:0ygD}
  \end{equation}
  However, substituting $y^*_i = z^*_i / w_i'$ into equation
  \eqref{eq:0ygD} immediately gives us the equivalence 
  \begin{equation}
    \Set{i \in \Vcal}{z^*_i - \beta'w'_i < 0} 
    \subseteq T^*(\beta')
    \subseteq \Set{i \in \Vcal}{z^*_i - \beta'w'_i \leq 0}.
    \label{eq:0ygD3}
  \end{equation}
  Letting $T^\dagger(\beta) = T^*(\beta / M)$, and recalling that
  $\beta' w'_i = \beta w_i$, we have that
  \begin{equation}
    U_1(\beta) = \Set{i \in \Vcal}{z^*_i - \beta w_i < 0} 
    \subseteq T^\dagger(\beta)
    \subseteq \Set{i \in \Vcal}{z^*_i - \beta w_i \leq 0} = U_2(\beta).
    \label{eq:0ygD4}
  \end{equation}
  Thus $U'_1(\beta') = U_1(\beta)$ and $U'_2(\beta') = U_2(\beta)$,
  and we have proved \ref{th:general-weights:1} for $\mw \in
  \Qbb^n_{++}$.

  $ $

  \nid Part \ref{th:general-weights:2} follows similarly. $\my^*$ in
  \eqref{eq:GNNt} is the minimum norm vector for the extended problem
  $f_{\mw'}$, so by theorem
  \ref{th:min-norm-sfm}-\ref{th:min-norm-sfm:2} $U_1'(\beta')$ and
  $U_2'(\beta')$ are the smallest and largest minimizers of $f(S) -
  \beta'\mw'(S) = f(S) - \beta\mw(S)$.  However, $U'_1(\beta') =
  U_1(\beta)$ and $U'_2(\beta') = U_2(\beta)$, so $U_1(\beta)$ and
  $U_2(\beta)$ satisfy part \ref{th:general-weights:2}. Thus we have
  proved the theorem for rational $\mw$.

  $ $

  \nid {\bf Step 2:} Now, it remains to show that this result extends to
  all real weights as well.  This is more difficult than it immediately
  seems, as $\mz^*$ depends on $\mw$ through an optimization over
  $B(f)$, and, unless $\Dcal = 2^{\Vcal}$, $B(f)$ is not necessarily
  bounded.  Thus it takes some care to show that the sets generated by
  the inequalities in \eqref{eq:33eo:1} and \eqref{eq:33eo:2} are the
  limit points corresponding to a sequence of rational $\mw$, and that
  they are the smallest and largest minimizers of $f(S) - \beta
  \mw(S)$ for all $\beta$.

  First, for convenience, define $f_{\mtlw,\beta} \ST \Dcal \mapsto \Reals$ as
  \begin{equation}
    f_{\mtlw,\beta} = f(S) - \beta \mtlw(S) 
    \label{eq:odHKv}
  \end{equation}
  and recall that this is a submodular function for any $\mtlw \in
  \Reals_{++}^n$. Let $\my^*(\mtlw,\beta)$ be the corresponding
  minimum norm vector:
  \begin{equation}
    \my^*(\mtlw,\beta) = \argmin_{\my \in B(f_{\mtlw, \beta})} \sqnorm{\my}.
    \label{eq:fAL81}
  \end{equation}
  Similarly, denote $\mz^*$ as a function of $\mtlw$ as well, i.e.
  \begin{equation}
    \mz^*(\mtlw) = \argmin_{\mz \in B(f)} \sum_{i \in \Vcal} \frac{z_i^2}{\tlw_i}.
    \label{eq:O0UXZ}
  \end{equation}
  Now suppose that $\mw \notin \Qbb^n$, so the above proof for
  rational $\mw$ does not apply. As $\Qbb$ is dense on $\Reals$ and
  $\mw > 0$, there exists a sequence of positive rationals $\momega_1,
  \momega_2, ...$, $\momega_i \in \Qbb^n$ such that
  \begin{equation}
    \lim_{i \goesto \infty} \momega_i = \mw. 
    \label{eq:DtjWT}
  \end{equation}
  By the minimum norm theorem and the fact that we have proved the
  theorem in question for all rational $\momega_m$, we know that, for
  all $i$ and $m$,
  \begin{align}
    U_1(\beta, \momega_m) 
    &= \Set{i}{z_i^*(\momega_m) - \beta \omega_{m,i} < 0} 
     = \Set{i}{y_i^*(\momega_m, \beta) < 0} \label{eq:5megh:1}\\
    U_2(\beta, \momega_m) 
    &= \Set{i}{z_i^*(\momega_m) - \beta \omega_{m,i} \leq 0} 
    = \Set{i}{y_i^*(\momega_m, \beta) \leq 0} \label{eq:5megh:2}. 
  \end{align}
  where we have made the dependence of the minimal and maximal sets
  $U_1(\beta)$ and $U_2(\beta)$ in \eqref{eq:33eo:1} and
  \eqref{eq:33eo:2} on $\momega_m$ explicit since we are working with
  a sequence of problems given by $\momega_m$.  Thus,
  by theorem \ref{th:min-norm-sfm}, for all $\mtlw \in \set{\mw, \momega_1, \momega_2, ...}$,
  \begin{align}
    U_1(\beta, \mtlw) &= \Set{i}{y_i^*(\mtlw, \beta) < 0} \\ 
   \intertext{and}
    U_2(\beta, \mtlw) &= \Set{i}{y_i^*(\mtlw, \beta) \leq 0} 
  \end{align}
  are the unique smallest and largest minimizing sets of $f_{\mtlw,
    \beta}$, respectively.  Thus we are done if we can show that, for
  all $\beta$,
  \begin{align}
    \lim_{m \goesto \infty} \quad\, \mz^*(\momega_m) - \beta \momega_m &= \mz^*(\mw) - \beta \mw \label{eq:rlw9i:1} \\
    \intertext{and}
    \lim_{m \goesto \infty} \my^*(\momega_m, \beta) &= \my^*(\mw, \beta) \label{eq:rlw9i:2},
  \end{align}
  as this immediately implies that \eqref{eq:5megh:1} and
  \eqref{eq:5megh:2} hold in the limit as well.  

  First, let $D$ be a convex domain, and consider the function $g_D^*\ST\Reals^n \times \Ioo{0,\infty}
  \mapsto D$, where
  \begin{equation}
    g^*_D(\mdelta, \mw) = \argmin_{\mz \in D}  \sum_{i = 1}^n \frac{(z_i + \delta_i)^2}{w_i}.
    \label{eq:0YDak}
  \end{equation}
  As $w_i > 0$ is fixed, we know that $g^*_D(\mdelta, \mw)$ is locally
  Lipshitz-continuous in $\mw$ by lemma \ref{lem:lipshitz}.  Thus, for
  all fixed $D$ and  sequences $(\mtldelta_m, \mtlw_m)$ such that
  \begin{equation}
    (\mtldelta_m, \mtlw_m) \goesto (\mdelta, \mw)  \quad \text{ as } \quad m \Goesto \infty,
    \label{eq:LAH23}
  \end{equation}
  we have that 
  \begin{equation}
    g^*_D(\mtldelta_m, \mtlw_m) \goesto g^*_D(\mdelta_m, \mw_m)  \quad \text{ as } \quad m \Goesto \infty.
    \label{eq:a6glf}
  \end{equation}
  Now, we may assume w.l.o.g. that $\momega_m > 0$.  Then we immediately have that 
  \begin{equation}
  \mz^*(\momega_m) = g^*_{B(f)}(\m0, \momega_m) \Goesto g^*_{B(f)}(\m0, \mw) = \mz^*(\mw) \quad \text{ as } \quad m \Goesto \infty,
    \label{eq:i4Ky7}
  \end{equation}
  proving \eqref{eq:rlw9i:1}.  

$ $

\nid To show \eqref{eq:rlw9i:2}, let $\mdelta_m = \momega_m - \mw$,
and note that
  \begin{align}
    B(f_{\momega_m, \beta}) 
    &= \Tcbrsss{\mx \in \Reals^n \ST
      \mx(S) \leq f(S) - \beta\momega_m(S) \Forall S \in \Dcal,\; \nonumber \\
      &\hspace{6.13em}
      \mx(\Vcal) = f(\Vcal) - \beta \momega_m(\Vcal)} \\
    &= \Tcbrsss{\mx + \beta\Tbr{\mw - \momega_m} \ST
      \mx(S) \leq f(S) - \beta\momega_m(S) - \beta\Tbr{\mw(S) - \momega_m(S)} 
      \Forall S \in \Dcal,\; \nonumber \\
      &\hspace{9.8em}
      \mx(\Vcal) = f(\Vcal) - \beta \momega_m(\Vcal) - \beta\Tbr{\mw(\Vcal) - \momega_m(\Vcal)}} \\
    &= \Tcbrsss{\mx + \beta\Tbr{\mw - \momega_m} \ST
      \mx(S) \leq f(S) - \beta\mw(S)   \Forall S \in \Dcal,\; \nonumber \\
      &\hspace{9.8em}
      \mx(\Vcal) = f(\Vcal) - \beta \mw(\Vcal)} \\
    &= \Set{\mx - \beta \mdelta_m}{\mx \in B(f_{\mw, \beta})}. 
    \label{eq:sHjuq}
  \end{align}
  Thus
  \begin{align}
    \my^*(\momega_m, \beta) 
    &= \argmin_{\my \in B(f_{\momega_m, \beta})} \sqnorm{\my} \\
    &= \argmin_{\my' \in B(f_{\mw, \beta})} \sqnorm{\my' - \beta \mdelta_m} \\
    &= g^*_{B(f_{\mw, \beta})}(\beta \mdelta_m, \m1).
  \end{align}
  Thus, since $\mdelta_m \goesto \m0$, 
  \begin{equation}
    \my^*(\momega_m, \beta) 
    = g^*_{B(f_{\mw, \beta})}(\beta \mdelta_m, \m1) 
    \Goesto g^*_{B(f_{\mw, \beta})}(\m0, \m1)
    = \my^*(\mw, \beta) \quad \text{ as } \quad m \Goesto \infty,
    \label{eq:jm4oa}
  \end{equation}
  proving \eqref{eq:rlw9i:2}.  The theorem is proved.
\end{proof}
%


\include{p1_proofs}







%
%
\bibliographystyle{plainnat}
\bibliography{references}
%
%
 
 


\end{document}